%% file: discomteval.tex
\documentclass{clv3}

\newif\ifdraft
\drafttrue

\usepackage{url}
\usepackage{latexsym}
\usepackage{graphicx}
\usepackage{booktabs}
\usepackage{subcaption}
\usepackage{multirow}
\usepackage{color}
\usepackage{amsmath,bm}
\usepackage{pgfplots}
\usepackage{tikz}
\usepackage{tkz-graph}
\usepackage{hyperref}
\usepackage{xcolor}
\definecolor{darkblue}{rgb}{0, 0, 0.5}
\hypersetup{colorlinks=true,citecolor=darkblue, linkcolor=darkblue, urlcolor=darkblue}

\input{sections/defs}

\issue{vv}{nn}{20yy}
\dochead{}
\runningtitle{Discourse Structure in Machine Translation Evaluation}
\runningauthor{Joty, Guzm\'an, M\`arquez and Nakov}

\historydates{\final{Submission received: August 26, 2016; revised version received: December 12, 2016; accepted for publication:  April 21, 2017}}

\begin{document}

\title{Discourse Structure\\ in Machine Translation Evaluation}

\author{Shafiq Joty\thanks{HBKU Research Complex B1, P.O. Box 5825, Doha, Qatar. E-mail: \{sjoty,fguzman,lmarquez,pnakov\}@qf.org.qa}}
\affil{Arabic Language Technologies (ALT), Qatar Computing Research Institute (QCRI), Hamad bin Khalifa University (HBKU), Qatar Foundation}

\author{Francisco Guzm\'an$^*$}
\affil{ALT, QCRI, HBKU, Qatar Foundation}

\author{Llu\'is M\`arquez$^*$}
\affil{ALT, QCRI, HBKU, Qatar Foundation}

\author{Preslav Nakov$^*$}
\affil{ALT, QCRI, HBKU, Qatar Foundation}

\maketitle

\begin{abstract}
In this article, we explore the potential of using sentence-level discourse structure for machine translation evaluation. We first design discourse-aware similarity measures, which use all-subtree kernels to compare discourse parse trees in accordance with the Rhetorical Structure Theory (RST). Then, we show that a simple linear combination with these measures can help improve various existing machine translation evaluation metrics regarding correlation with human judgments both at the segment- and at the system-level. This suggests that discourse information is complementary to the information used by many of the existing evaluation metrics, and thus it could be taken into account when developing richer evaluation metrics, such as the WMT-14 winning combined metric \discoparty. We also provide a detailed analysis of the relevance of various discourse elements and relations from the RST parse trees for machine translation evaluation. In particular, we show that (i)~all aspects of the RST tree are relevant, (ii)~nuclearity is more useful than relation type, and (iii)~the similarity of the translation RST tree to the reference RST tree is positively correlated with translation quality.
\end{abstract}


\section{Introduction}
\label{sec:intro}
\input{sections/introduction}

\section{Discourse-Based Similarity Measures}
\label{sec:measures}
\input{sections/measures}

\section{Experimental Setting}
\label{sec:setting}

\input{sections/setting}

\section{Evaluation of the Discourse-based Metrics}
\label{sec:evaluation}
\input{sections/ACL-WMT-exp}

\section{Analysis}
\label{sec:analysis}
\input{sections/analysis}


\section{Related Work}
\label{sec:related}
\input{sections/related}
\section{Conclusions}
\label{sec:conclusions}
\input{sections/conclusions}

\section*{Acknowledgments}
The authors would like to thank the anonymous reviewers for their constructive and thorough comments and criticism. Thanks to their comments we have improved this work significantly. 

\bibliographystyle{compling}
\bibliography{CL}

\end{document}

%% file: sections/defs.tex
\DeclareMathOperator*{\Sigm}{Sig}

\newcommand{\final}[1]{\textcolor{black}{#1}}

\newcommand{\qcri}{\nobreak{DR}}
\newcommand{\qcril}{\nobreak{DR-{\sc lex}}}

\newcommand{\sqcri}{\nobreak{\sc dr}}
\newcommand{\sqcril}{\nobreak{\sc dr-{\scriptsize lex}}}
\newcommand{\full}{\nobreak{\emph{full}}}
\newcommand{\nor}{\nobreak{\emph{no\_rel}}}
\newcommand{\non}{\nobreak{\emph{no\_nuc}}}
\newcommand{\nonr}{\nobreak{\emph{no\_nuc \& no\_rel}}}
\newcommand{\nostr}{\nobreak{\emph{no\_discourse}}}

\newcommand{\fo}{\nobreak{F$_1$}}
\newcommand{\rmse}{\nobreak{RMSE}}

\newcommand{\dummy}{\nobreak{\emph{dummy}\hspace{3pt}}}

\newcommand{\drLEX}{\nobreak{DR-{\sc lex}$_{1}$}}
\newcommand{\dr}{\nobreak{DR-{\sc nolex}}}

\newcommand{\drLEXoo}{\nobreak{DR-{\sc lex}$_{1.1}$}}

\newcommand{\disco}{\nobreak{DiscoTK}}

\newcommand{\discoparty}{\nobreak{{\sc DiscoTK}$_{party}$}}

\newcommand{\spede}{\nobreak{\sc spede07pP}}
\newcommand{\sempos}{\nobreak{SEMPOS}}
\newcommand{\amber}{\nobreak{AMBER}}
\newcommand{\meteor}{\nobreak{\sc Meteor}}
\newcommand{\terrorcat}{\nobreak{\sc TerrorCat}}
\newcommand{\simpbleu}{\nobreak{SIMPBLEU}}
\newcommand{\ter}{\nobreak{TER}}
\newcommand{\bleu}{\nobreak{BLEU}}
\newcommand{\posf}{\nobreak{\sc posF}}
\newcommand{\block}{\nobreak{\sc BlockErrCats}}

\newcommand{\word}{\nobreak{\sc WordBlockEC}}

\newcommand{\xen}{\nobreak{\sc XEnErrCats}}

\newcommand{\nist}{\nobreak{NIST}}
\newcommand{\rouge}{\nobreak{\sc Rouge}}

\newcommand{\asiya}{\nobreak{\sc Asiya}}

\newcommand{\csen}{\nobreak{\sc cs-en}}
\newcommand{\deen}{\nobreak{\sc de-en}}
\newcommand{\esen}{\nobreak{\sc es-en}}
\newcommand{\fren}{\nobreak{\sc fr-en}}
\newcommand{\ruen}{\nobreak{\sc ru-en}}
\newcommand{\hien}{\nobreak{\sc hi-en}}

\newcommand{\wmtf}{\nobreak{\sc wmt14}}

\newcommand{\Ni}{({\em i})~}
\newcommand{\Nii}{({\em ii})~}
\newcommand{\Niii}{({\em iii})~}
\newcommand{\Niv}{({\em iv})~}

%% file: sections/introduction.tex
%


From its foundations, Statistical Machine Translation (SMT) as a field had two defining characteristics:
first, translation was modeled as a generative process at the \emph{sentence-level}.
Second, it was purely statistical over words or word sequences and made little to no use 
of \emph{linguistic information}~\cite{brown93mathematic,Koehn:2003:SPT}.  

Although modern SMT systems switched to a discriminative log-linear framework \cite{och03minimum,watanabe-EtAl:2007,Chiang:2008,Hopkins2011}, which allows for additional sources as features, it is generally 
hard to incorporate dependencies beyond a small window of adjacent words, thus making it difficult to use linguistically-rich models.

\noindent One of the fruitful research directions for improving SMT has been the usage of more structured linguistic information. For instance, in SMT we find systems based on syntax~\cite{GalleyHKM04,Quirk:2005}, hierarchical structures~\cite{Chiang:2005:HPM}, and semantic roles~\cite{wu-fung:2009:NAACLHLT09-Short,Lo2012,bazrafshan-gildea:2014:EMNLP2014}. 
However, it was not until recently that syntax-based SMT systems started to outperform their phrase-based counterparts, especially for language pairs that need long-distance reordering such as Chinese-English and German-English~\cite{nadejde-williams-koehn:2013:WMT}.
%

Another less explored way consists of going beyond the sentence-level, for example, translating at the document level or taking into account broader contextual information. The idea is to obtain adequate translations respecting cross-sentence relations, and enforcing cohesion and consistency at the document level~\cite{Hardmeier12,ben-EtAl:2013:Short,louis-webber:2014:EACL,tu-zhou-zong:2014:P14-1,Xiong:2015}.
Research in this direction has also been the focus of the two editions of the DiscoMT workshop, in 2013 and 2015 \cite{DiscoMT2013,DiscoMT2015,DiscoMT2015:sharedtask}.

\emph{Automatic MT evaluation} is an integral part of the process of developing and tuning an SMT system.
Reference-based evaluation measures compare the output of a system to one or more human translations (called \emph{references}) and produce a similarity score indicating the quality of the translation. The first metrics approached similarity as a shallow word $n$-gram matching between the translation and one or more references, with a limited use of linguistic information. BLEU~\cite{Papineni:Roukos:Ward:Zhu:2002} is the best-known metric in this family, which has been used for years as the evaluation standard in the MT community. 
BLEU can be efficiently calculated and has shown good correlation with human assessments when evaluating systems on large quantities of text. However, it is also known that BLEU and similar 
 metrics are unreliable for high-quality translation output~\cite{Doddington:2002:AEM,lavie-agarwal:2007:WMT}, and they cannot tell apart raw machine translation output from a fully fluent professionally post-edited version thereof~\cite{denkowski2012challenges}. Moreover, lexical-matching similarity has been shown to be both insufficient and not strictly necessary for two sentences to convey the same meaning~\cite{Culy:2003,Coughlin:2003,Callison-Burch:2006}.

Several alternatives emerged to overcome these limitations, 
most notably TER~\cite{Snover06astudy} and METEOR~\cite{Lavie:2009:MMA}. Researchers have explored, with good results, the addition of other levels of linguistic information including synonymy and paraphrasing \cite{Lavie:2009:MMA}, syntax~\cite{Gimenez2007,Popovic2007,Liu2005}, semantic roles \cite{Gimenez2007,Lo2012}, and, most recently, discourse~\cite{Comelles2010,Wong2012,guzman-EtAl:ACL2014,discoMT:WMT2014,guzman-EtAl:2014:EMNLP2014}.


On top of all previous considerations, MT systems are usually evaluated by computing translation quality on individual sentences and performing some simple aggregation to produce the \emph{system-level} evaluation scores. To the best of our knowledge, semantic relations between clauses in a sentence and between sentences in a text have not been seriously explored. However, clauses and sentences rarely stand on their own in a well-written text; rather, the logical relationship between them carries significant information that allows the text to express a meaning as a whole. Each clause 
follows smoothly from the ones before it and leads into the ones that come afterward. This logical relationship between clauses forms a \emph{coherence structure} \cite{Hobbs79}. In discourse analysis, we seek to uncover this coherence structure underneath the text.


\noindent Several formal theories of discourse have been proposed to describe the coherence structure~\cite{Mann88,Alex03,Webber04}. Rhetorical Structure Theory or RST \cite{Mann88} is perhaps the most influential of them in computational linguistics, where it is used either to parse the text in language understanding or to plan a coherent text in language generation \cite{Taboada06}. RST describes coherence using  \emph{discourse relations} between parts of a text and postulates a hierarchical tree structure called \emph{discourse tree}. For example, Figure \ref{fig:DTs} shows discourse trees for three different translations of a source sentence.


Modeling discourse brings together the usage of higher-level linguistic information and the exploration of relations between clauses and sentences in a text, which makes it a very attractive goal for MT and its evaluation. We believe that the semantic and pragmatic information captured in the form of discourse trees \Ni can yield better MT evaluation metrics, and \Nii can help develop discourse-aware SMT systems that produce more coherent translations.

In this work, we focus on the first of the two previous research hypotheses. 
%
Specifically, we show that sentence-level discourse information can be used to produce reference-based evaluation measures  that perform well on their own, but more importantly, can be used to improve over many existing MT evaluation metrics regarding correlation with human assessments. We conduct our research in three steps.
First, we design a simple discourse-aware similarity measure, \qcril, based on RST trees, generated with a publicly-available discourse parser~\cite{jotycodra:15}, and the well-known \emph{all sub-tree} kernel~\cite{Collins01}. The sub-tree kernel computes a similarity value by comparing the discourse tree representation of a system translation with that of a reference translation. We show that a simple uniform linear combination with this metric helps to improve a large number of MT evaluation metrics at the segment-level and at the system-level in the context of the WMT11 and the WMT12 metrics shared task benchmarks~\cite{WMT11,WMT12}.
Second, we show that tuning (i.e., learning) the weights in the linear combination of metrics using human assessed examples is a robust way to improve the effectiveness of the \qcril\ metric significantly. 
Following the idea of an interpolated combination, we put together several variants of our discourse metric (using different tree-based representations) with many strong pre-existing metrics provided by the \asiya\ toolkit for MT evaluation~\cite{gonzalez-gimenez-marquez:2012:Demo}. The result is \discoparty, which scored best at the WMT14 Metrics task~\cite{WMT14}, both at the system and at the segment levels.
Third, we conduct an ablation study which helps us understand which elements of the discourse parse tree have the highest impact on the quality of the evaluation measure. Interestingly enough, the \emph{nuclearity} feature (i.e., the distinction between main and subordinate units) of the RST tree turns out to be more important than the discourse relation types (e.g., \emph{Elaboration, Contrast}).

Note that although extensive, this study is restricted to sentence-level evaluation, which arguably can limit the benefits of using global discourse properties, i.e., document-level discourse structure. Fortunately, many sentences are long and complex enough to present rich discourse structures connecting their basic clauses.
Thus, although limited, this setting can demonstrate the potential of discourse-level information for MT evaluation.
Furthermore, sentence-level scoring is compatible with most translation systems, which work on a sentence-by-sentence basis. It could also be beneficial to modern MT tuning mechanisms such as PRO 
\cite{Hopkins:2011} and MIRA \cite{watanabe-EtAl:2007,Chiang:2008}, which also work at the sentence-level. Finally, it could also be used for re-ranking $n$-best lists of translation hypotheses.


%

\noindent The rest of the paper is organized as follows.
Section~\ref{sec:measures} introduces our proposal for a family of discourse-based similarity metrics.
Sections~\ref{sec:setting} and~\ref{sec:evaluation} describe the experimental setting and the evaluation of the discourse-based metrics, alone and in combination with other pre-existing measures.
Section~\ref{sec:analysis} empirically analyzes the main discourse-based metric and performs an ablation study to better understand its contributions.
%
%
Finally, Sections~\ref{sec:related} and~\ref{sec:conclusions} discuss related work and present the conclusions together with some directions for future research.

%% file: sections/measures.tex
%

Different formal theories of discourse have been proposed in the literature, reflecting different viewpoints about how it is best to describe the coherence structure of a text. For example, \namecite{Alex03} proposed the Segmented Discourse Representation Theory (SDRT), which is driven by sentence semantics. \namecite{Webber04} and \namecite{Danlos09} extended the sentence grammar to formalize discourse structure. \namecite{Mann88} proposed the Rhetorical Structural Theory (RST), which was inspired by empirical analysis of authentic texts. Although RST was initially intended to be used for text generation, it later became popular as a framework for parsing the structure of a text. This work relies on RST-based coherence structure.

\begin{figure*}[t]
  \begin{subfigure}{.8\linewidth}
  \mbox{\hspace*{-0.5cm}
  \includegraphics[scale=0.65]{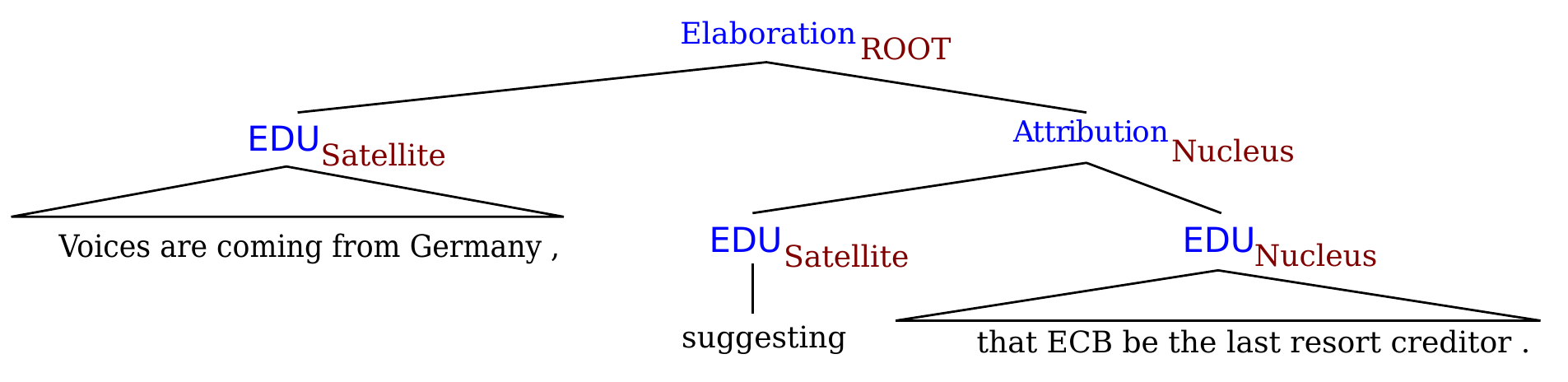}}
  \caption{\bf A reference (human) translation.}
  \label{fig:gold}
  \end{subfigure}

  \begin{subfigure}{.8\linewidth}
  \mbox{\hspace{0.2cm}
     \includegraphics [scale=0.7] {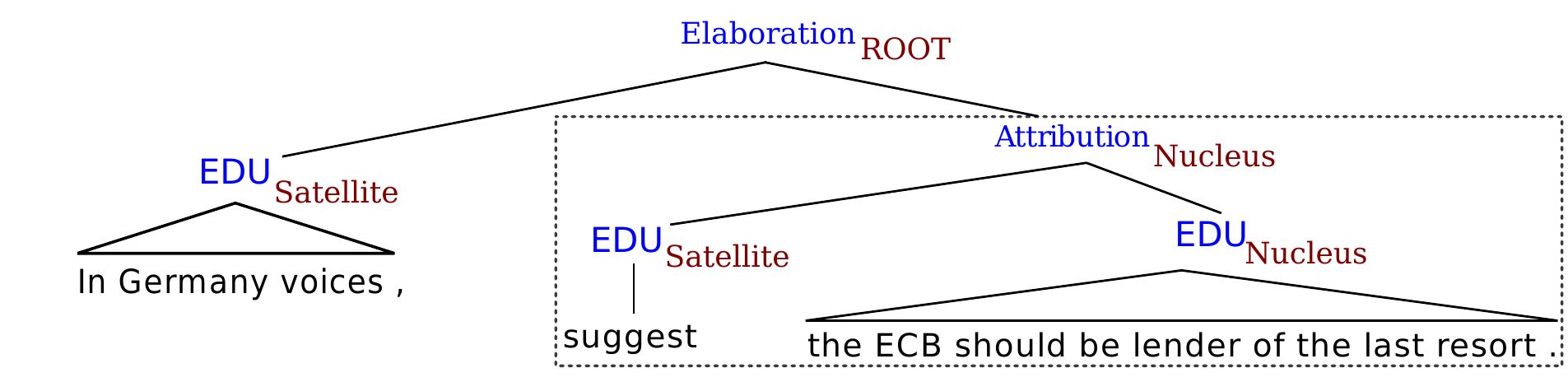}}
      \caption{\bf A higher quality system translation.}
       \label{fig:sys1}
  \end{subfigure}

  \begin{subfigure}{.8\linewidth}
  \mbox{\hspace{2cm}
     \includegraphics [scale=0.8] {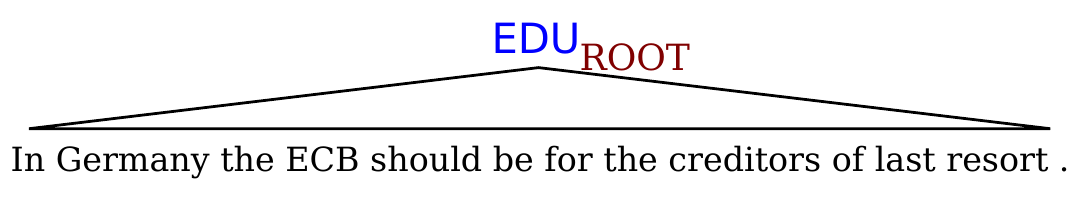}}
      \caption{\bf A lower quality system translation.}
       \label{fig:sys2}
  \end{subfigure}
  \caption{\label{fig:DTs}Example of three different discourse trees for the translations of a source sentence: (a) the reference, (b) a higher-quality translation, (c) a lower-quality translation.}
  \vspace{2mm}
\end{figure*}

RST posits a tree representation of a text, known as a discourse tree. As shown in Figure~\ref{fig:gold},  the leaves of a discourse tree (three in this example) correspond to contiguous atomic clause-like text spans, called \textbf{Elementary Discourse Units (EDUs)}, which serve as building blocks for constructing the tree. In the tree, adjacent EDUs are connected by certain \textbf{coherence relations} (e.g., \emph{Elaboration}, \emph{Attribution}), thus forming larger discourse units, which in turn are also subject to this process of relation-based linking. Discourse units that are linked by a relation are further distinguished based on their relative importance in the text: \textbf{nuclei} are the core arguments of the relation, while \textbf{satellites} are supportive ones. A discourse relation can be either \textbf{mononuclear} or  \textbf{multinuclear}. A mononuclear relation connects a nucleus and a satellite (e.g., \emph{Elaboration}, \emph{Attribution} in Figure~\ref{fig:gold}), where a multinuclear relation connects two or more nuclei (e.g., \emph{Joint}, \emph{Contrast}). Thus, a RST discourse tree comprises four types of elements: \Ni EDUs that comprise textual information, \Nii the structure or skeleton of the tree, \Niii nuclearity statuses of the discourse units, and \Niv coherence relations by which adjacent discourse units are linked.

Our hypothesis in this article is that the similarity between the discourse trees of an automatic and of a reference translation provides additional information that can be valuable for evaluating MT systems. In particular, we believe that better system translations should be more similar to the human translations in their discourse structures than worse ones. As an example, consider the three discourse trees shown in Figure~\ref{fig:DTs}: (\emph{a}) for a reference translation, and (\emph{b}) and (\emph{c}) for translations of two different systems from the WMT12 competition. Notice that the tree structure, the nuclearity statuses and the relation labels in the reference translation are also realized in the system translation in (b), but not in (c); this makes (b) a better translation compared to (c), according to our hypothesis. We argue that existing metrics that only use lexical and syntactic information cannot distinguish well between the translations in (b) and (c).
     
In order to develop a discourse-aware evaluation metric, we first generate discourse trees for the reference and the system-translated sentences using an RST discourse parser, and then we measure the similarity between the two trees. We describe these two steps in more detail below.

\subsection{Generating Discourse Trees} \label{sec: gen_dt}

Conventionally, discourse analysis in RST involves two main subtasks: \Ni {\it discourse segmentation}, or breaking the text into a sequence of EDUs, and \Nii{\it discourse parsing}, or the task of linking the discourse units (which could be EDUs or larger units) into labeled discourse trees. Recently, \namecite{Shafiq12,jotycodra:15} proposed discriminative models for discourse segmentation and discourse parsing. Their discourse segmenter uses a maximum entropy model and achieves state-of-the-art performance with an F$_1$-score of 90.5, while human agreement for this task is 98.3 in F$_1$-score.

The discourse parser uses a dynamic Conditional Random Field~\cite{Sutton07} as a parsing model to infer the probability of all possible discourse tree constituents. The inferred (posterior) probabilities are then used in a probabilistic CKY-like bottom-up parsing algorithm to find the most likely parse. Using the standard set of 18 coarse-grained discourse relations,\footnote{See \cite{Carlson01} for a detailed description of the discourse relations.} the discourse parser achieved an $F_1$-score of 79.8\% at the sentence level, which is close to the human agreement of 83\%. These high numbers inspired us to develop discourse-aware MT evaluation metrics.\footnote{A demo of the parser is available at \url{http://alt.qcri.org/demos/Discourse_Parser_Demo/} The source code of the parser is available from \url{http://alt.qcri.org/tools/discourse-parser/}}

\subsection{Measuring Similarity between Discourse Trees}

A number of metrics have been proposed to measure the similarity between two labeled trees, e.g., Tree Edit Distance \cite{Tai79} and various Tree Kernels (TKs) \cite{Collins01,Alex-NIPS03,Moschitti-ECML2006}. One advantage of tree kernels is that they provide an effective way to integrate tree structures in kernel-based learning algorithms like SVMs, and learn from arbitrary tree fragments as features. 

\begin{figure*}[t]
  \centering
  \includegraphics[width=5.3in,height=2.8in]{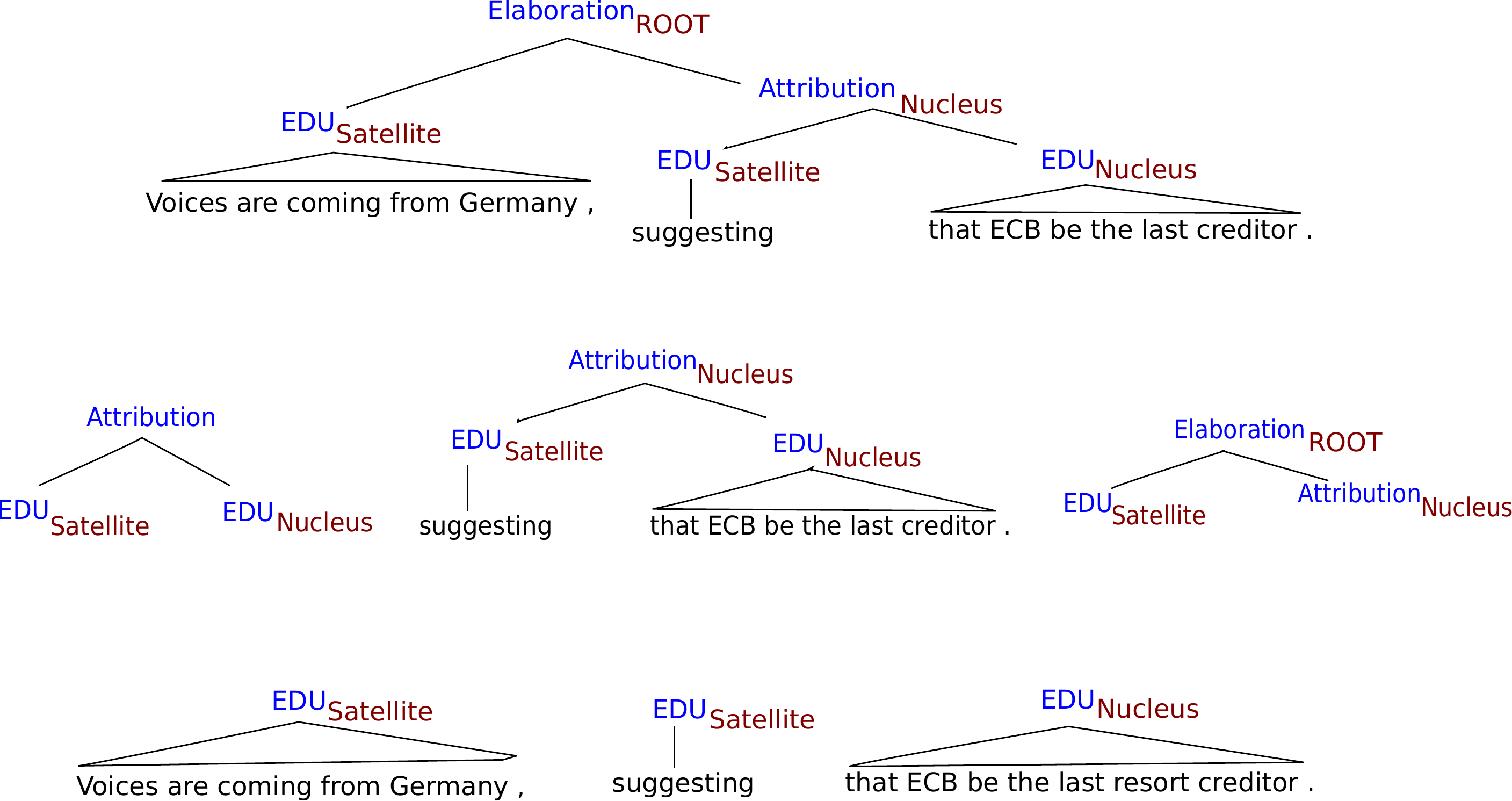}
  \caption{Discourse subtrees used by the syntactic tree kernel for the tree in Figure~\ref{fig:gold}.}
  \label{fig:tree_frags}
\end{figure*}

\namecite{Collins01} proposed a syntactic tree kernel to efficiently compute the number of common subtrees in two syntactic trees. To comply with the rules (or productions) of a context-free grammar in syntactic parsing, the subtrees in this kernel are subject to the constraint that their nodes are taken with either all or none of the children. Since the same constraint applies to discourse trees, we use the same tree kernel in our work. Figure \ref{fig:tree_frags} shows the valid subtrees according to the syntactic tree kernel for the discourse tree in Figure \ref{fig:gold}. Note that in this work we use the tree kernel only to measure the similarity between two discourse trees rather than to learn subtree features in a supervised kernel-based learning framework like SVM. As an example of the latter, see our more recent work \cite{guzman-EtAl:2014:EMNLP2014}, which uses tree kernels over syntactic and discourse structures in an SVM preference ranking framework. 

\namecite{Collins01} proposed two modifications of the kernel when using it in a classifier (e.g., SVM) to avoid the classifier behaving like a nearest neighbour rule: \Ni to restrict the tree fragments considered in the kernel computation based on their depth, and/or \Nii to assign relative weights to the tree fragments based on their size. Since we do not use the kernel in a learning algorithm, these modifications do not apply to us; all subtrees are equally weighted in our kernel.



Figure \ref{fig:tree_frags} shows that, when applied to discourse trees, the syntactic tree kernel may limit us on the type of substructures that we wish to compare. For example, although matching the complete production (i.e., a parent with all of its children) may make more sense for subtrees with internal nodes only (i.e., non-terminals), we may want to relax this constraint at the terminal (text) level to allow word subsequence matches. 

\noindent One way to cope with this limitation of the tree kernel is to change the representation of the trees to a form that is suitable to capture the relevant information for our task. \final{For example, in order to allow for the syntactic tree kernel to find subtree matches at the word unigram level, we can include an artificial layer of leaves, e.g., by copying the same \emph{dummy} label below each word. In this way, the words become pre-terminal nodes and can be matched against the words in the other tree.}



Apart from the above modification to match subtrees at the word level, we experimented with different representations of a discourse tree, each of which produces a different discourse-based evaluation metric. 
In this section we present two basic representations of the discourse tree, namely \qcri\ and \qcril, which we will use in Section \ref{sec:evaluation} to demonstrate that the discourse measures are synergetic 
with several 
widely used MT evaluation metrics (DR stands for discourse representation).

\begin{figure*}[t]
 \begin{subfigure}[t]{0.8\linewidth}
 \centering
  \includegraphics[scale=0.9]{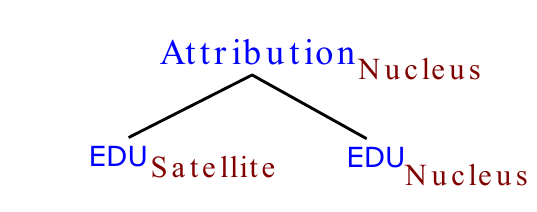}
  \caption{Discourse tree representation for \qcri.}
  \label{fig:dr}
  \end{subfigure}
  
 \begin{subfigure}[t]{.8\linewidth}
 \centering
  \includegraphics[scale=0.77]{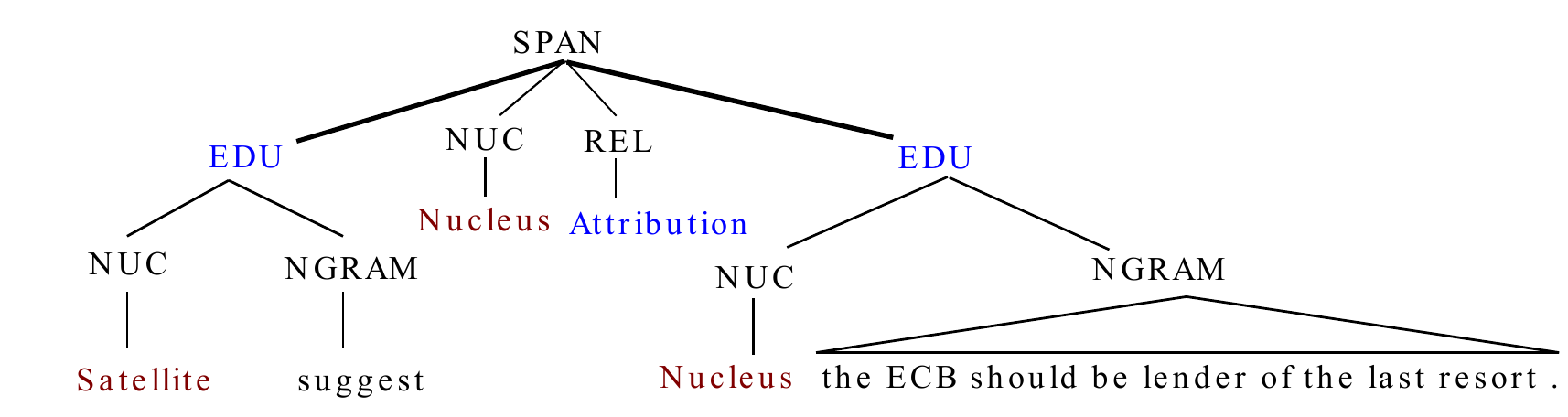}
  \caption{Discourse tree representation for \qcril}
  \label{fig:drl}
  \end{subfigure}	
  \caption{Two discourse tree representations for the highlighted subtree in Figure \ref{fig:sys1}.}
  \label{fig:reps}
\end{figure*}

Figure~\ref{fig:reps} shows the two representations \qcri \ and \qcril\ for the highlighted subtree in Figure \ref{fig:sys1}, that spans the text: \emph{``suggest the ECB should be the lender of last resort''}. As shown in Figure \ref{fig:dr}, \qcri \ does not include any lexical item. Therefore, the syntactic tree kernel, when applied to this representation of the discourse tree, measures the similarity between two candidate translations in terms of their discourse representations only. 

On the contrary, \qcril, as shown in Figure \ref{fig:drl}, includes the lexical items to account for lexical matching; moreover, it separates the structure (skeleton) of the tree from its labels, i.e., the nuclearity statuses and the relation labels. This allows the syntactic tree kernel to give partial credit to subtrees that differ in labels but match in their skeletons or vice versa. More specifically, \qcril \ uses the predefined tags SPAN and EDU to build the skeleton of the tree, and considers the nuclearity and/or the relation labels as properties, added as children, of these tags. For example, a SPAN has two properties: its nuclearity status and its relation label, where an EDU has only one property: its nuclearity status. The words of an EDU are placed under another predefined tag NGRAM. \final{To allow the tree kernel to find subtree matches at the word level, we also include an additional layer of \emph{dummy} leaves (for simplicity, not shown in Figure \ref{fig:drl})}.


%% file: sections/setting.tex
In this section, we describe the datasets we used in our experiments,  the interpolation approach we applied to combine our discourse-based metrics with pre-existing evaluation metrics, and all the correlation measures we used for evaluation.

\subsection{Datasets}

In our experiments, we used the data available for the WMT11, WMT12, WMT13, and WMT14 metrics shared tasks for translations into English.\footnote{\final{\url{http://www.statmt.org/wmtYY/results.html}, with YY in $\{11,12,13,14\}$.}}
This includes the output from the systems that participated in the MT evaluation campaigns in those four years and the corresponding English reference translations.
The WMT11 and WMT12 datasets contain 2,000 and 3,003 sentences, respectively, for each of the following four language pairs: Czech-English (\csen), French-English (\fren), German-English (\deen), and Spanish-English (\esen). In WMT13, the Russian-English (\ruen) pair was added to the mix, and the dataset has 3,000 sentences for each of the five language pairs. WMT14 dropped \esen \ and included Hindi-English (\hien), with each language pair having 3,003 sentences, except for \hien, for which there were 2,507 sentences. 

The task organizers provided human judgments on the quality of the systems' translations.
These judgments represent rankings of the output of five systems chosen at random, for a particular language-pair and for a particular sentence.
The overall coverage, i.e., the number of unique sentences that were evaluated, was only a fraction of the total (see Table \ref{tab:wmt-stats}). For example, for WMT11 \fren, only 247 out of 3,000 sentences have human judgments.
While the evaluation setup of WMT evaluation is performed in a sentence-level fashion, we believe that it is adequate for our purpose. The annotation interface allowed human judges to take  longer-range discourse structure into account, as they were shown the source and the human reference translations in the context of one preceding and one following sentences.\footnote{\final{A detailed description of the WMT evaluation setting can be found in~\cite{WMT14}.}}


%


\begin{table}[t]
\centering
\small
\begin{tabular}{lr@{ }@{ }@{ }c@{ }@{ }@{ }c@{ }@{ }@{ }cr@{ }@{ }@{ }c@{ }@{ }@{ }c@{ }@{ }@{ }cr@{ }@{ }@{ }c@{ }@{ }@{ }c@{ }@{ }@{ }cr@{ }@{ }@{ }c@{ }@{ }@{ }c@{ }@{ }@{ }c}
&\multicolumn{3}{c}{\bf{WMT11}}&&\multicolumn{3}{c}{\bf{WMT12}} && \multicolumn{3}{c}{\bf{WMT13}}&& \multicolumn{3}{c}{\bf{WMT14}}\\
&  sys &  pairs &  sent && sys & pairs & sent && sys & pairs & sent && sys & pairs & sent \\
    \cmidrule{2-4}\cmidrule{6-8}\cmidrule{10-12}\cmidrule{14-16}
  \bf \csen & 12 & 2,477 & 190 &&  6 & 8,269 & 937 && 11 & 46,397 & 2,572 &&  5 & 10,301 & 1,288\\
  \bf \deen & 28 & 7,358 & 346 && 16 & 9,084 & 968 && 17 & 75,856 & 2,589 && 13 & 15,971 & 1,472\\
  \bf \esen & 21 & 4,799 & 274 && 12 & 8,751 & 910 && 12 & 36,626 & 2,172 && -- & --     & --\\
  \bf \fren & 24 & 5,085 & 247 && 15 & 8,747 & 932 && 13 & 43,234 & 2,272 &&  8 & 15,033 & 1,365\\
  \bf \ruen & -- & --    & --  && -- & --    & --  && 19 & 94,509 & 2,740 && 13 & 24,595 & 1,800\\
  \bf \hien & -- & --    & --  && -- & --    & --  && -- & --     & --    &&  9 & 14,678 & 1,180\\
 \bottomrule
\end{tabular}
\caption{\label{tab:wmt-stats} Number of systems (sys), unique non-tied translation pairs (pairs), and unique sentences for which such pairs exist (sent) for the different language pairs, for the human evaluation of the WMT11-WMT14 metric shared tasks. These statistics show what we use for \emph{training}; the numbers for \emph{testing} are higher, as explained in the text.}
\end{table}

Table~\ref{tab:wmt-stats} shows the main statistics about the data that we used for \emph{training}, where we excluded all pairs for which: \Ni both translations were judged as equally good, or \Nii the number of votes for translation$_1$ being better than translation$_2$ equals the number of votes for it being worse than translation$_2$.
Moreover, we ignored repetitions, i.e., if two judges voted the same way, we did not create two training examples, but just one (note, however, that on testing, repetitions will be accounted for).
Excluding ties and repetitions reduces the number of training pairs significantly, e.g., for WMT13 \csen, we have 46,397 pairs, while initially there were 85,469 judgments in total. 

Note, however, that for \emph{testing}, we used the official full datasets, where we used all pairwise judgments, including judgments saying that both translations are equally good, ties in the number of wins of translation$_1$ vs. translation$_2$ and repetitions. This is important to make our results fully comparable to previously-published work.

As a final analysis on the WMT corpora, we studied the complexity of the discourse trees. Recall that we imposed the limitation of working with discourse structures at the sentence level. If we want the discourse metrics to be impactful, we need to make sure that a significant number of sentences have non-trivial discourse trees.

\begin{figure}[t]
\centering
\pgfplotsset{width=7.5cm,compat=1.3}
\begin{tikzpicture}
	\begin{axis}[
		xlabel=Depth of discourse tree,
		yticklabel style={/pgf/number format/fixed},
		ylabel=Proportion of sentences]

	\addplot[color=blue,mark=x] coordinates {
(0,0.28)
(1,0.2445)
(2,0.199)
(3,0.119)
(4,0.077)
(5,0.033)
(6,0.021)
(7,0.012)
(8,0.0075)
(9,0.0035)
(10,0.002)
	};
	
	\addplot[color=red,mark=o] coordinates {

(0,0.255744255744256)
(1,0.26007326007326)
(2,0.2001332001332)
(3,0.122544122544123)
(4,0.0715950715950716)
(5,0.0406260406260406)
(6,0.0243090243090243)
(7,0.010989010989011)
(8,0.00599400599400599)
(9,0.00366300366300366)
(10,0.00166500166500167)
	};

	\addplot[color=green,mark=triangle] coordinates {

(0,0.291)
(1,0.268333333333333)
(2,0.197)
(3,0.116666666666667)
(4,0.0616666666666667)
(5,0.032)
(6,0.0183333333333333)
(7,0.00633333333333333)
(8,0.003)
(9,0.00466666666666667)
(10,0.000333333333333333)
	};
	
\legend{WMT11, WMT12, WMT13}

	\end{axis}
\end{tikzpicture}
\caption{Distribution of sentences by tree depth computed based on the reference translations of WMT11, WMT12 and WMT13.}
\label{fig:sent-freq-vs-depth}
\end{figure}
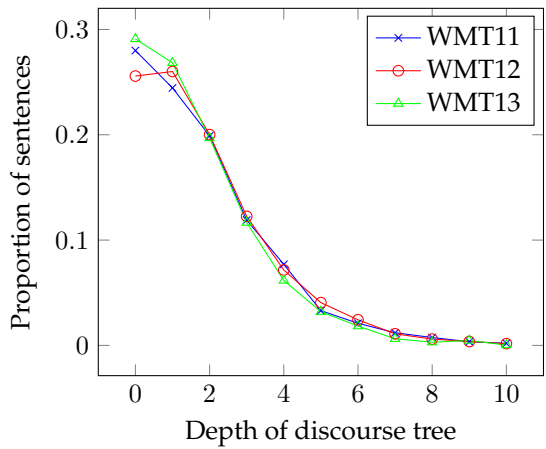

Figure~\ref{fig:sent-freq-vs-depth} shows the proportion of sentences by discourse tree depth for the WMT11, WMT12 and WMT13 datasets. We computed these statistics with our automatic discourse parser applied to the reference translations. As it can be seen, the three datasets show very similar curves. One relevant observation is that more than 70\% of the sentences have a non-trivial discourse tree (depth $>0$). Of course, the proportion of sentences decreases quickly with the tree depth. About 20\% of the sentences have trees of depth 2 and slightly over 10\% have trees of depth 3. The average depth for the three datasets is 1.77, with a minimum absolute value of 0 and a maximum of 32. The number of EDUs contained in those trees average to 2.77, with a minimum number of 1 and a maximum number of 33. 
Although the impact of discourse information is potentially higher at the paragraph or document level, we showed that we have complex enough sentences in our datasets in terms of discourse structure. Thus, we have justified that there is potential in testing the effect of discourse information in MT evaluation metrics.

%

\subsection{Learning Interpolation Weights for Metric Combination}
\label{DataPreprocess}

In Subsections~\ref{subsec:segment-level} and \ref{subsec:discotk}, we report experiments with a simple linear model combining the predictions of our discourse-based metrics with several pre-existing MT evaluation metrics. The interpolation weights are trained discriminatively from the manually annotated rankings described in the previous subsection. In order to use the annotations for training, we first transformed the five-way relative rankings into ten pairwise comparisons.
For instance, if a judge has ranked the output of systems $A$, $B$, $C$, $D$, $E$ as ${A>B>C>D>E}$,
this would entail the following ten pairwise rankings:
${A>B}$, ${A>C}$, ${A>D}$, ${A>E}$,
${B>C}$, ${B>D}$, ${B>E}$,
${C>D}$, ${C>E}$,
and
${D>E}$.
We then use a \emph{maximum entropy} learning framework to learn the interpolation weights, where  the classification task is to distinguish a better translation from a worse one for each pair of translation hypotheses. The log likelihood of the training data with $l_2$ regularization on the weight parameters $\mathbf{w}$ is:\vspace*{-5mm}

\begin{equation}
J(\mathbf{w}) = \sum_{i=1}^N y_i\log~\Sigm \left(\mathbf{w}^T (\mathbf{u_i^1} - \mathbf{u_i^2})\right) + (1-y_i)\log\left(1-\Sigm (\mathbf{w}^T (\mathbf{u_i^1} - \mathbf{u_i^2}))\right) + \lambda \mathbf{w}^T\mathbf{w}
\end{equation}

\noindent where $\Sigm(x)$ is the \emph{sigmoid} (aka \emph{logistic}) function, the $\mathbf{u_i^1}$ and $\mathbf{u_i^2}$ vectors represent the values of the evaluation measures we are combining for the two translations in the pair $i = (t_1, t_2)$, and  $y_i\in\{1,0\}$ is the human assessment for the pair, i.e., $y_i = 1$ if $t_1$ is better than $t_2$, otherwise $y_i = 0$.  We learn the model parameters $\mathbf{w}$ using the L-BFGS fitting algorithm, which is time- and space-efficient. We learn the regularization strength parameter $\lambda$ using $5$-fold cross-validation on the training dataset.\footnote{When fitting the model, we did not include a bias term, as this was harmful.}

Note that our approach to learn the interpolation weights is similar to the one used by PRO for tuning the relative weights of the components of a log-linear SMT model~\cite{Hopkins:2011}. Unlike PRO,
\Ni \ we used \emph{human judgments}, not automatic scores, and
\Nii \ we trained on \emph{all pairs}, not on a subsample.

\subsection{Correlation Measures}
\label{Corelation}

In our experiments, we only considered translation into English (as we had a discourse parser for English only), and we used the data described in Table~\ref{tab:wmt-stats}. For evaluation, we followed the standard setup of the Metrics task of WMT12~\cite{WMT12}. 
For segment-level evaluation,
we used \emph{Kendall's $\tau$} \cite{Kendall}, which can be calculated directly from the human pairwise judgments.
For system-level evaluation, we used \emph{Spearman's rank correlation} \cite{Spearman} and, in some cases, also \emph{Pearson correlation} \cite{Pearson}, which are appropriate correlation measures as here we have vectors of scores.

We measured the correlation of the evaluation metrics with the human judgments provided by the task organizers. As we explained above, the judgments represent rankings of the output of five systems chosen at random, for a particular sentence, also chosen at random. From each of those rankings, we produce ten pairwise judgments (see \ref{DataPreprocess} above). Then, using those pairwise human judgments, we evaluated the performance of the different 
MT evaluation metrics at the segment or at the system level.

\subsubsection{Segment-level Evaluation}
We used Kendall's $\tau$ to measure the correlations between the segment-level scores\footnote{In this section we use the term \emph{segment} instead of \emph{sentence} as we do in the rest of the paper, to be consistent with the terminology used in the MT field.} given by a target evaluation metric and the human judgments.
Kendall's $\tau$ is defined as follows:
\begin{equation}
\tau = \frac{\#concordant - \#discordant}{\#concordant + \#discordant}
\end{equation}
\noindent where $\#concordant$ is the number of concordant translation pairs, i.e., for which the human ranking and the corresponding metric scores agree, and $\#discordant$ is the number of pairs for which the human ranking and the metric score disagree.
For example, if the human judgment is that the translation of system $s_i$ for segment $k$ is better than the translation of system $s_j$ for segment $k$, this pair will be considered concordant if the metric gives higher score to $s_i$ than to $s_j$ for segment $k$.

The value of Kendall's $\tau$ ranges between -1 (all pairs are discordant) and 1 (all pairs are concordant), and negative values are worse than positive ones.
Note that different sets of systems may be ranked for the different segments, but in the calculations, we only use pairs of systems for which we have human judgments. Such direct judgments are available for a particular language pair and for a particular segment. We do not calculate Kendall's $\tau$ for each language pair; instead, we consider all pairwise judgments as part of a single set (as implemented in the official WMT scripts).

In the original Kendall's $\tau$ \cite{Kendall}, comparisons with human or metric ties are considered neither concordant nor discordant. In the experiments in Section~\ref{sec:evaluation}, we used the official scorers from the WMT Metrics tasks to compute Kendall's $\tau$. More precisely, in Subsections~\ref{subsec:system-level} and \ref{subsec:segment-level} we use the WMT12
  version of Kendall's $\tau$ \cite{WMT12}, while in Subsection~\ref{subsec:discotk} we report results using the WMT14 scorer \cite{machacek-bojar:2014:W14-33}.
 We used these two different versions of the software to allow a direct comparison the official results that were reported for the metrics task in WMT12  \cite{WMT12} and WMT14 \cite{machacek-bojar:2014:W14-33}.

\subsubsection{System-level Evaluation}

For the correlation at the system level, we first produce a score for each of the systems according to the quality of their translations based on the evaluation metrics and on the human judgments. Then, we calculate the correlation between the scores for the participating systems using a target metric's scores and the human scores. We do this based on system's ranks induced by the scores (using \emph{Spearman's rank correlation}) or based on the scores themselves (using \emph{Pearson correlation}).
Note that, following WMT, we calculate the correlation score separately for each language pair, and then we average the resulting correlations to obtain the final score.

\paragraph{Segment-to-system score aggregation} In order to produce a system-level score based on the pairwise sentence-level human judgments, we need to aggregate these judgments, which we do based on the ratio of wins (ignoring ties), as defined for the official ranking at WMT12 \cite{WMT12}:\footnote{We use the WMT12 aggregation script. See also \cite{WMT12} for a discussion and comparison of several aggregation alternatives.}
\begin{equation}
\mathrm{score}(s_i) = \frac{\mathrm{win}(s_i)}{\mathrm{win}(s_i) + \mathrm{loss}(s_i)}
\end{equation}
\noindent where $\mathrm{win}(s_i)$ and $\mathrm{loss}(s_i)$ are the number of wins and losses, respectively, of system $s_i$ against any other system in the segment-level pairwise human judgments.


\paragraph{Spearman's rank correlation} This is the WMT12 official metric for system-level evaluation. To calculate it, we first convert the raw scores assigned to each system to ranks, and then we use the following formula \cite{Spearman}:
\begin{equation}
\rho = 1 - \frac{6 \sum{d_{i}^2}}{n(n^2-1)}
\end{equation}
\noindent where $d_i$ is the difference between the ranks for system $i$, and $n$ is the number of systems being evaluated.
Note that this formula requires that there be no ties in the ranks of the systems (based on the automatic metric or based on the human judgments), which was indeed the case. Spearman's rank correlation ranges between -1 and 1. However, unlike Kendall's $\tau$, here the sign does not matter, and high \emph{absolute} values indicate better performance. In our experiments, we used the official script from WMT12. 

In some experiments, we also report \emph{Pearson correlation} \cite{Pearson}, which was the official system-level score at WMT14.
This is a more general correlation coefficient than Spearman's and does not require that all $n$ ranks be distinct integers. It is defined as follows:
\begin{equation}
r = \frac{\sum_{i=1}^n{(H_i - \bar{H}) (M_i - \bar{M})}}{\sqrt{\sum_{i=1}^n{(H_i - \bar{H})^2}} \sqrt{\sum_{i=1}^n{(M_i - \bar{M})^2}}}
\end{equation}
\noindent where $H$ is the vector of the human scores of all participating systems, $M$ is the vector of the corresponding scores as predicted by the given metric, and $\bar{H}$ and $\bar{M}$ are the means for $H$ and $M$, respectively.

The Pearson correlation value ranges between -1 and 1, where higher absolute score is better. We used the official WMT14 scoring tool to calculate it.

%% file: sections/ACL-WMT-exp.tex
%

In this section, we show the utility of discourse information for machine translation evaluation. Below we present the evaluation results at the system- and at the segment-level, using our two basic discourse-based metrics, which we
refer to as \qcri\ and \qcril \ (Subsection \ref{sec: gen_dt}).
In our experiments, we combine \qcri\ and  \hbox{\qcril} with other evaluation metrics in two different ways: using uniform linear interpolation (at the system- and at the segment-level), and using a tuned linear interpolation for the segment-level. We only present the average results over all language pairs. For clarity, in our tables we show results divided into three evaluation groups:

\paragraph{Group I} contains our discourse-based evaluation metrics, \qcri \ and \qcril.
\paragraph{Group II} includes the publicly available MT evaluation metrics that participated in the WMT12 metrics task, excluding those that did not have results for all language pairs~\cite{WMT12}. More precisely, they are \spede,  \amber, \meteor, \terrorcat, \simpbleu, \xen, \word, \block, and \posf.
\paragraph{Group III} contains other important individual evaluation metrics that are commonly used in MT evaluation: \bleu~\cite{Papineni:Roukos:Ward:Zhu:2002}, \nist~\cite{Doddington:2002:AEM}, 
\rouge~\cite{Lin04}, and \ter~\cite{Snover06astudy}.
We calculated the metrics in this group using Asiya. In particular, we used the following Asiya versions of \ter\ and \rouge: \textsc{TERp-A} and \textsc{ROUGE-w}.\footnote{\final{These variants are described in page 19 of the \asiya\ manual 
(\url{http://asiya.lsi.upc.edu/).}}}
\medskip

For each metric in groups II and III, we present the system-level and segment-level results for the original metric as well as for the linear interpolation of that metric with \qcri\ and with \qcril. The combinations with \qcri \ and \qcril \ that improve over the original metrics are
shown in bold, and those that yield degradation are in \emph{italic}.

For the segment-level evaluation, we further indicate which interpolated results yield statistically significant improvement over the original metric. Note that testing statistical significance is not trivial in our case since we have a complex correlation score for which the assumptions that standard tests make are not met. We thus resorted to 
a non-parametric randomization framework \cite{Yeh:2000:MAT}, which is commonly used in NLP research.\footnote{We did not apply the significance test at the system level due to the insufficient number of scores available to sample from: there were a total of 49 (language pair, system) scores for the WMT12 data.}

\subsection{System-level Results}
\label{subsec:system-level}


Table~\ref{tab:results-wmt12-sys} shows the system-level experimental results for WMT12. We can see that \qcri\ is already competitive by itself: on average, it has a correlation of .807, which is very close to the BLEU and the TER scores from group II (.810 and .812, respectively). Moreover, \qcri \ yields improvements when combined with 13 of the 15 metrics, with a resulting correlation higher than those of the two individual metrics being combined. This fact suggests that \qcri \ contains information that is complementary to that used by most of the other metrics.

\input{tables/tab-results-wmt12-sys}

As expected, \qcril\ performs better than \qcri\ since it is lexicalized (at the unigram level), and also gives partial credit to correct structures. Individually, \qcril\ outperforms most of the metrics from group II, and ranks as the second best metric in that group.  Furthermore, when combined with individual metrics, \qcril \ is able to improve 14 out of the 15 metrics. 
Averaging over all metrics in the table, the combination of  \qcri\ improves the average of the individual metrics correlation from .816 to .852 (+.035) and \qcril\ further improves the average results up to .868 (+.052). 

Thus, we can conclude that at the system level, adding discourse information to a metric, even using the simplest of the combination schemes, 
is a good idea for most of the metrics.



\subsection{Segment-level Results} 
\label{subsec:segment-level}

Table~\ref{tab:results-wmt12-seg-notuned-tuned} shows the results for WMT12 at the segment-level.
We can see that \qcri \ performs badly, with a high negative Kendall's $\tau$  of -.433.
This should not be surprising because \Ni the discourse tree structure alone does not contain enough information for a good evaluation at the segment level, and \Nii this metric is more sensitive to the quality of the DT, which can be wrong or void.
Moreover, \qcri\ is more likely to produce a high number of ties, which is harshly penalized by WMT12's definition of Kendall's $\tau$.
Conversely, ties and incomplete discourse analysis were not a problem at the system level, where evidence from all 3,003 test sentences is aggregated, allowing us to rank systems more precisely.
Due to the low score of \qcri \ as an individual metric, it fails to yield improvements when uniformly combined with other metrics (see ``Untuned +\qcri'' column).

\input{tables/tab-results-wmt12-seg-notuned-tuned}

Again, \qcril \ is better than \qcri; with a positive $\tau$ of .133, yet as an individual metric, it ranks poorly compared to other metrics in groups II and III.
However, when uniformly combined (see ``Untuned +\qcri'' column) with other metrics, \qcril \ outperforms 9 of the 13 metrics in Table~\ref{tab:results-wmt12-seg-notuned-tuned}, with statistically significant improvements in 8 of these cases ($p$-value <.01).


Following the learning method described in Subsection \ref{DataPreprocess}, we experimented also with tuning the interpolation weights
in the metric combinations.  We report results for \Ni cross-validation on WMT12, and \Nii tuning on WMT12 and testing on WMT11.

\paragraph{Cross-validation on WMT12}
For cross-validation on WMT12, we used ten folds of approximately equal sizes, each containing about 300 sentences: we constructed the folds by putting together entire documents, thus not allowing sentences from a document to be split over two different folds. During each cross-validation run, we trained our pairwise ranker using the human judgments corresponding to nine of the ten folds. We then used the remaining fold for evaluation. Note that in this process, we aggregated the data for different language pairs, and we produced a single set of tuning weights for all language pairs.\footnote{Tuning separately for each language pair yielded slightly lower results.}

The results are shown in the last two columns of Table~\ref{tab:results-wmt12-seg-notuned-tuned} (``Tuned'').
We can see that the tuned combinations with \qcril\ improve over all but one of the individual metrics in groups II and III, with statistically significant differences in 10 out of the 12 cases. Even more interestingly, the tuned combinations that include the much weaker metric \qcri\ now improve over 12 out of 13 of the individual metrics, being 9 of this differences statistically significant with $p$-value <.01. 
This is remarkable given that \qcri\ has a strong negative $\tau$ as an individual metric at the sentence-level. Again, these results suggest that both \qcri\ and \qcril\ contain information that is complementary to that of the individual metrics that we experimented with.

Averaging over all 13 cases, \qcri\ improves Kendall's $\tau$ from .159 to .193 (+.035), while \qcril\ improves it to .211 (+.053). These sizable improvements highlight the importance of tuning the linear combination when working at the segment level.


\paragraph{Testing on WMT11}
To rule out the possibility that the improvement of the tuned metrics on WMT12 could have come from over-fitting, and also in order to verify that the tuned metrics generalize when applied to other sentences, we also tested on an additional dataset: WMT11.
We tuned the weights for our metric combinations on \emph{all} WMT12 pairwise judgments (no cross-validation), and we evaluated them on the WMT11 dataset. Since the metrics that participated in WMT11 and WMT12 are different (and even when they have the same name, there is no guarantee that they have not changed from 2011 to 2012), this time we only report results for the standard Group III metrics, thus ensuring that the metrics in the experiments are consistent for 2011 and 2012.

The results, presented in Table~\ref{tab:results-wmt11-seg-tuned}, show the same pattern as before:
\Ni adding \qcri\ and \qcril\ improve over all individual metrics, being the differences statistically significant in 7 out of the 8 cases with $p$-value <.01; and \Nii the contribution of \qcril\ is consistently larger than that of \qcri.
Observe that these improvements are very close to those for the WMT12 cross-validation. This shows that the weights learned on WMT12 generalize well, as they are also good for WMT11.

\input{tables/tab-results-wmt11-seg-tuned}
\subsection{DR-based Metrics in a Strong MT Evaluation Measure}
\label{subsec:discotk}

From the results presented in the previous subsections, we can conclude that discourse structure is an important information source, which is not entirely correlated to other information sources considered so far, and thus should be taken into account when designing future metrics for automatic evaluation of machine translation output. In this subsection we show how the simple combination of DR-based metrics with a selection of other existing strong MT evaluation metrics can lead to a very competitive evaluation metric, \discoparty~\cite{discoMT:WMT2014}, which we presented at the metrics task of WMT14~\cite{machacek-bojar:2014:W14-33}.

\asiya\ \cite{AsiyaPBML:2010} is a suite for MT evaluation that provides a large set of metrics using different levels of linguistic
information. We used the 12 individual metrics from  \asiya's ULC~\cite{AsiyaMTjournal:2010}, which was the
best performing metric both at the system- and at the segment-level at the WMT08 and WMT09 metrics tasks. 
From the original ULC, we  replaced Meteor by the four newer variants \meteor-ex (exact match), \meteor-st (+stemming),  \meteor-sy (+synonymy lookup) and \meteor-pa (+paraphrasing) in \asiya's terminology  \cite{Denkowski2011}. We also added to the mix  \ter p-A (a variant of \ter\ with paraphrasing), \bleu, \nist, and \rouge-W, for a total of 18 individual metrics. The metrics in this set use diverse linguistic information, including lexical-, syntactic- and semantic-oriented individual metrics.

Regarding the discourse metrics, we used 5 variants, including \qcri\ and \qcril\ described in Section~\ref{sec:measures}, and three more constrained variants oriented to match words between trees only if they occur under the same substructure types (e.g., the same nuclearity type). These variants are designed by introducing structural  modifications in the discourse trees. A detailed description can be found in~\cite{discoMT:WMT2014}.


We tuned the relative weights of the previous 
23 individual metrics (18~\asiya + 5~discourse) following the same maximum entropy learning framework described in Subsection~\ref{DataPreprocess}. As the training set, we used the simple concatenation of WMT11, WMT12, and WMT13.

\begin{table*}[t]
\small
\begin{center}
\begin{tabular}{@{ }lcccccc@{ }}
\bf System & \fren & \deen & \hien & \csen & \ruen & Overall\\
\hline
\discoparty   & \bf 0.433$^{**}$ & \bf 0.380$^{**}$ & 0.434      & \bf 0.328$^{**}$ & \bf 0.355$^{**}$ &\bf 0.386$^{**}$\\
Best at \wmtf & 0.417\phantom{$^{**}$}      & 0.345\phantom{$^{**}$}     & \bf 0.438  & 0.284\phantom{$^{**}$}   & 0.336\phantom{$^{**}$}  & 0.364\phantom{$^{**}$}  \\
              & +.016\phantom{$^{**}$}               & +.035 \phantom{$^{**}$}              & -.004              & +.044\phantom{$^{**}$}  & +.019\phantom{$^{**}$}  & +.024\phantom{$^{**}$} \\
\hline
\end{tabular}
\end{center}
\caption{\label{tab:WMT14:comparetothebest:segment}Comparing our tuned metric to the best rivaling metric at \wmtf, for each individual language pair (this best rival differs across language pairs) at the segment-level using Kendall's $\tau$. Statistically significant improvements are marked with $^{**}$ for $p$-value < 0.01.}

\end{table*}


\discoparty\ was the best-performing metric at WMT14 both at the segment and at the system level, among a set of 16 and 20 participants, respectively ~\cite{machacek-bojar:2014:W14-33}. Table~\ref{tab:WMT14:comparetothebest:segment} shows a comparison at the segment level of our tuned metric \discoparty\ to the best rivaling metric at \wmtf, for each individual language pair, using Kendall's $\tau$. 
Note that this best rival differs across language pairs, e.g., for \fren, \hien, and \csen\ it is \textsc{beer}, for \deen\ it is \textsc{upc-stout}, and for \ruen\ it is \textsc{REDcombSent}. We can see that our metric outperforms this best rival for four of the language pairs, with statistically significant differences. 
The only exception is \hien, where the best rival performs slightly better, not being this difference statistically significant.

System translations for Hindi-English 
\final{were of extremely low quality}, and were very hard to discourse-parse accurately.\footnote{Note that the Hindi-English language pair caused a similar problem for a number of other metrics at the \wmtf\ Shared Task competition that relied on linguistic analysis.} The linguistically-heavy components of our \discoparty\ (discourse parsing, syntactic parsing, semantic role labeling, etc.) may suffer from the common ungrammaticality of the translation hypotheses for \hien, while other, less linguistically-heavy metrics seem to be more robust in such cases.

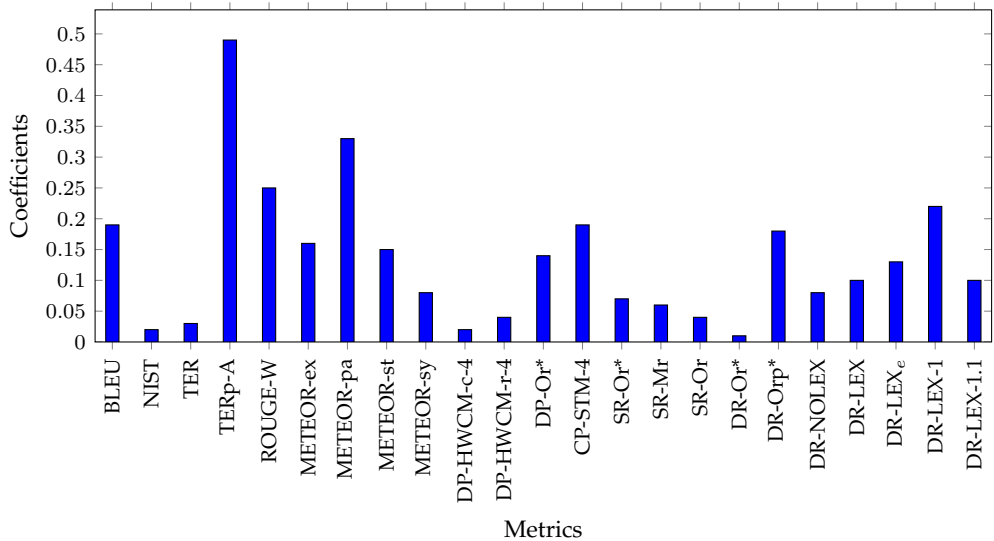
\begin{figure}[t!]
\centering
\hspace*{-5mm}
\pgfplotsset{width=9.8cm,compat=1.3}
\begin{tikzpicture}
\begin{axis}[
footnotesize,
symbolic x coords={BLEU, NIST, TER, TERp-A, ROUGE-W, METEOR-ex, METEOR-pa, METEOR-st, METEOR-sy, DP-HWCM-c-4, DP-HWCM-r-4, DP-Or*, CP-STM-4, SR-Or*, SR-Mr, SR-Or,DR-Or*,DR-Orp*, DR-NOLEX, DR-LEX, DR-LEX$_e$, DR-LEX-1,DR-LEX-1.1},
ybar=1pt, 
bar width=5pt,
ymin=0,   
width=13.5cm,
enlarge x limits=0.02,
height=6cm,
xlabel={Metrics},
ylabel={Coefficients},
x tick label style={rotate=90,anchor=east, font=\footnotesize},
yticklabel style={/pgf/number format/fixed},
xtick=data]

\addplot [fill=blue]
coordinates {(BLEU, 0.19) (NIST, 0.02) (TER, 0.03) (TERp-A, 0.49) (ROUGE-W, 0.25) (METEOR-ex, 0.16) (METEOR-pa, 0.33) (METEOR-st, 0.15) (METEOR-sy, 0.08) (DP-HWCM-c-4, 0.02) (DP-HWCM-r-4, 0.04) (DP-Or*, 0.14) (CP-STM-4, 0.19) (SR-Or*, 0.07) (SR-Mr, 0.06) (SR-Or, 0.04) (DR-Or*, 0.01) (DR-Orp*, 0.18) (DR-NOLEX, 0.08) (DR-LEX, 0.10) (DR-LEX-1, 0.22)  (DR-LEX-1.1, 0.10) (DR-LEX$_e$, 0.13)};

\end{axis}
\end{tikzpicture}
\caption{Absolute coefficient values after tuning the \discoparty\ metric on the WMT11+12+13 dataset.}
\label{fig:freq}
\end{figure}

We show in Figure~\ref{fig:freq} the weights for the individual metrics combined in \discoparty\ after tuning on the combined WMT11+12+13 dataset. The horizontal axis displays all the individual metrics involved in the combination. The first block of metrics (from BLEU to DR-Orp*) consists of the 18 \asiya\ metrics. The last five (from \dr \ to \drLEXoo) metrics are the metric variants based on discourse trees. Note that all metric scores are passed through a \emph{min-max} normalization step to put them in the same scale before tuning their relative weights.
 
We can see that most of the metrics involved in the metric combination play a significant role, the most important ones being TERp-A, METEOR-pa (paraphrases), and ROUGE-W. Some metrics accounting for syntactic and semantic information also get assigned relatively high weights (\hbox{DP-Or*}, CP-STM-4, and DR-Orp*). 
Interestingly, all five variants of our discourse metric received moderately-high weights, with the four variants using lexical information (\qcril 's) being more important. In particular, \drLEX\ has the fourth highest absolute weight in the overall combination. This confirms again the importance of discourse information in machine translation evaluation.

%% file: tables/tab-results-wmt12-sys.tex
\begin{table}[t]
\begin{center}
{\small\begin{tabular}{@{}l@{}rrr@{ }@{ }r}
&\bf Metrics &  & +\sqcri & +\sqcril\\
\toprule
\multirow{ 2}{*}{\bf I}&\qcri  & .807 & -- & -- \\
&\qcril & .876 & -- & -- \\
\midrule
\multirow{ 11}{*}{\bf II}&\sempos & .902 & \it .853 & \bf .903 \\
&\amber & .857 & \it .829 & \bf .869 \\
&\meteor & .834 & \bf .861 & \bf .888 \\
&\terrorcat & .831 & \bf .854 & \bf .889 \\
&\simpbleu & .823 & \bf .826 & \bf .859 \\
&\ter & .812 & \bf .836 & \bf .848 \\
&\bleu & .810 & \bf .830 & \bf .846 \\
&\posf & .754 & \bf .841 & \bf .857 \\
&\block & .751 & \bf .859 & \bf .855 \\
&\word & .738 & \bf .822 & \bf .843 \\
&\xen & .735 & \bf .819 & \bf .843 \\
\midrule
\multirow{5}{*}{\bf III}&\bleu & .791 & \bf .880 & \bf .859 \\
&\nist & .817 & \bf .842 & \bf .875 \\
&\rouge & .884 & \bf .899 & \it .869 \\
&\ter & .908 & \bf .926 & \bf .920 \\
\bottomrule
\end{tabular}}
\end{center}
\caption{\label{tab:results-wmt12-sys}Results on WMT12 at the system-level (calculated on 6 systems for \csen, 16 for \deen, 12 for \esen, and 15 for \fren). Spearman's correlation with human judgments.}
\vspace{-5mm}
\end{table}

%% file: tables/tab-results-wmt12-seg-notuned-tuned.tex
\begin{table}[t]
\begin{center}
{\small\begin{tabular}{l@{ }rrr@{ }@{ }rr@{ }@{ }r}
& & & \multicolumn{2}{c}{\bf Untuned} & \multicolumn{2}{c}{\bf Tuned}\\
\cmidrule{4-5}\cmidrule{6-7}
&\bf Metrics &  \bf Orig. & +\sqcri & +\sqcril & +\sqcri & +\sqcril\\
\toprule
\multirow{ 2}{*}{\bf I}&\qcri & -.433 & -- & -- & -- & --\\
&\qcril & .133 & -- & -- & -- & --\\
\midrule
\multirow{ 9}{*}{\bf II}&\spede & .254 & \it .190 & \it .223\phantom{$^{**}$} & \it .253\phantom{$^{**}$} &  .254\phantom{$^{**}$}\\
&\meteor & .247 & \it .178 & \it .217\phantom{$^{**}$} & \bf .250\phantom{$^{**}$}& \bf .251\phantom{$^{**}$}\\
&\amber & .229 & \it .180 & \it .216\phantom{$^{**}$} & \bf .230\phantom{$^{**}$} & \bf .232\phantom{$^{**}$} \\
&\simpbleu & .172 & \it .141 & \bf .191$^{**}$ & \bf .181$^{**}$ & \bf .199$^{**}$ \\
&\xen & .165 & \it .132 & \bf .185$^{**}$ & \bf .175$^{**}$ & \bf .194$^{**}$ \\
&\posf & .154 & \it .125 & \bf .201$^{**}$ & \bf .160$^{**}$ & \bf .201$^{**}$ \\
&\word & .153 & \it .122 & \bf .181$^{**}$ & \bf .161$^{**}$ & \bf .189$^{**}$ \\
&\block & .074 & \it .068 & \bf .151$^{**}$ & \bf .087$^{**}$ & \bf .150$^{**}$ \\
&\terrorcat & -.186 & \bf -.111 & \bf -.104$^{**}$ & \bf .181$^{**}$ & \bf .196$^{**}$ \\
\midrule
\multirow{5}{*}{\bf III}&\bleu & .185 & \it .154 & \bf .190\phantom{$^{**}$} & \bf .189\phantom{$^{**}$} & \bf .194$^{*}$\phantom{$^{*}$}\\
&\nist & .214 & \it .172 & \it .206\phantom{$^{**}$} & \bf .222$^{**}$ & \bf .224$^{**}$ \\
&\rouge & .185 & \it .144 & \bf .201$^{**}$ & \bf .196$^{**}$ & \bf .218$^{**}$ \\
&\ter & .217 & \it .179 & \bf .229$^{**}$ & \bf .229$^{**}$ & \bf .246$^{**}$ \\
\bottomrule
\end{tabular}}
\end{center}
\caption{\label{tab:results-wmt12-seg-notuned-tuned}Results on WMT12 at the segment-level (calculated on 11,021 pairs for \csen, 11,934 for \deen, 9,796 for \esen, and 11,594 for \fren): untuned and tuned versions. Kendall's $\tau$ with human judgments. Improvements over the baseline are shown in bold, and statistically significant improvements are marked with $^{**}$ and $^{*}$ for $p$-value < 0.01 and $p$-value < 0.05, respectively.}
\vspace{-5mm}

\end{table}

%% file: tables/tab-results-wmt11-seg-tuned.tex
\begin{table}[t]
\begin{center}
{\small\begin{tabular}{@{}l@{ }@{ }rrc@{ }@{ }c}

& & & \multicolumn{2}{c}{\bf Tuned}\\
\cmidrule{4-5}
&\bf Metrics &  \bf Orig. & +\sqcri & +\sqcril\\
\toprule
\multirow{ 2}{*}{\bf I}&\qcri & -.447 & -- & --\\
&\qcril & .146 & -- & -- \\
\midrule
\multirow{5}{*}{\bf III}&\bleu & .186 & \bf .192\phantom{$^{**}$} & \bf .207$^{**}$ \\
&\nist & .219 & \bf .226$^{**}$ & \bf .232$^{**}$ \\
&\rouge & .205 & \bf .218$^{**}$ & \bf .242$^{**}$ \\
&\ter & .262 & \bf .274$^{**}$ & \bf .296$^{**}$ \\
\bottomrule
\end{tabular}}
\end{center}
\caption{\label{tab:results-wmt11-seg-tuned}Results on WMT11 at the segment-level (calculated on 3,695 pairs for \csen, 8,950 for \deen, 5,974 for \esen, and 6,337 for \fren): tuning on the entire WMT12.
         Kendall's $\tau$ with human judgments. Improvements over the baseline are shown in bold, and statistically significant improvements are marked with $^{**}$ for $p$-value < 0.01.}
 \vspace{-3mm}
\end{table}

%% file: sections/analysis.tex
%

When dealing with evaluation metrics based on lexical matching, such as \bleu\ or \nist, it is easier to understand how and why they work, and what their limitations are. However, if a metric deals with complex structures like discourse trees, it is not straightforward to explain its performance. 

In this section, we aim to understand better which parts of the discourse trees have the biggest impact on the performance of the discourse-based measures presented in Section~ \ref{sec:measures}. For that purpose, we first conduct an ablation study (see Subsection \ref{subsec:ablation}), where we dissect the different components of the discourse trees, and we analyze the impact that the deletion of such components has on the performance of our evaluation metrics. In a second study (see Subsection \ref{subsec:relations}), we analyze which parts of a complete discourse tree are most useful to distinguish between good and bad translations.
Overall, the components that we focus on in our analysis are the following: 
\Ni\emph{Discourse relations}, e.g., Elaboration, Attribution, etc.;
\Nii\emph{Nuclearity statuses}, i.e., Nucleus and Satellite;
\Niii\emph{Discourse structure}, e.g., boundaries of the elementary discourse units, depth of the tree, etc.

%
%

The previous two studies focus on quantitative aspects of the discourse trees. Subsection~\ref{subsec:qualitative} discusses one real example to understand from a more qualitative point of view the contribution of the sentence-level discourse trees in the evaluation of good and bad translations.
Finally, in Subsection~\ref{subsec:beyondsyntax}, we discuss the issue of whether discourse trees provide information that is complementary to syntax.

\subsection{Ablation Study at the System Level}
\label{subsec:ablation}
We analyze the performance of our discourse-based metric \qcril\ at the system level. We use \qcril\  instead of \qcri, as it exhibits the most competitive performance, and incorporates both lexical and discourse information. We selected system-level evaluation because the metric is much more stable and accurate at the system level than at the segment level. 

\begin{figure}[t]
\centering
\begin{tabular}{cl}
\includegraphics[width=.7\textwidth]{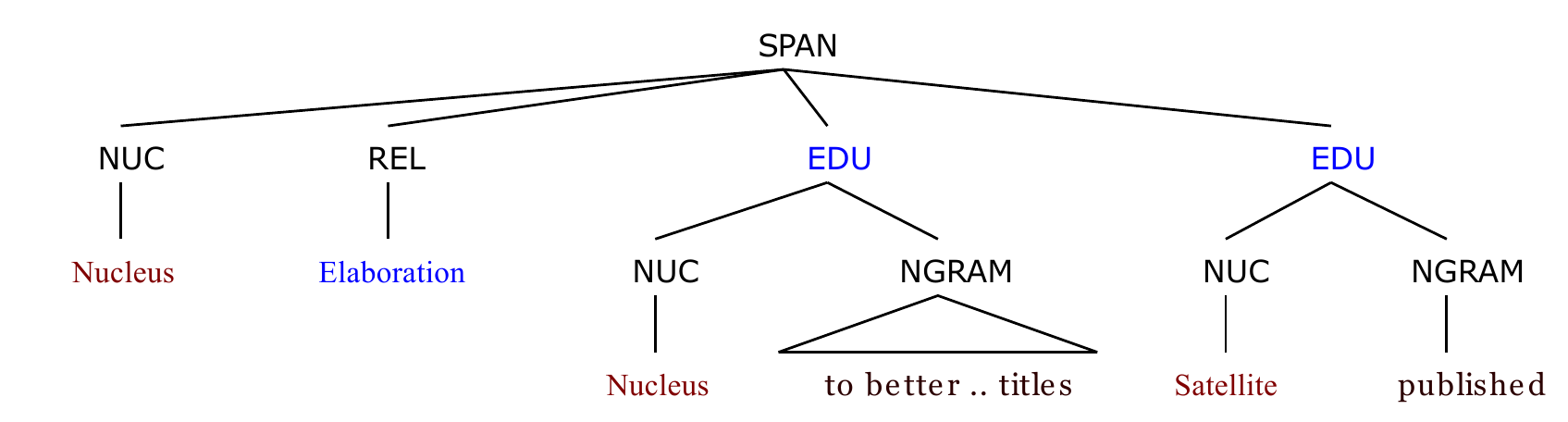} & {\small (1) \full}\\
\includegraphics[width=.7\textwidth]{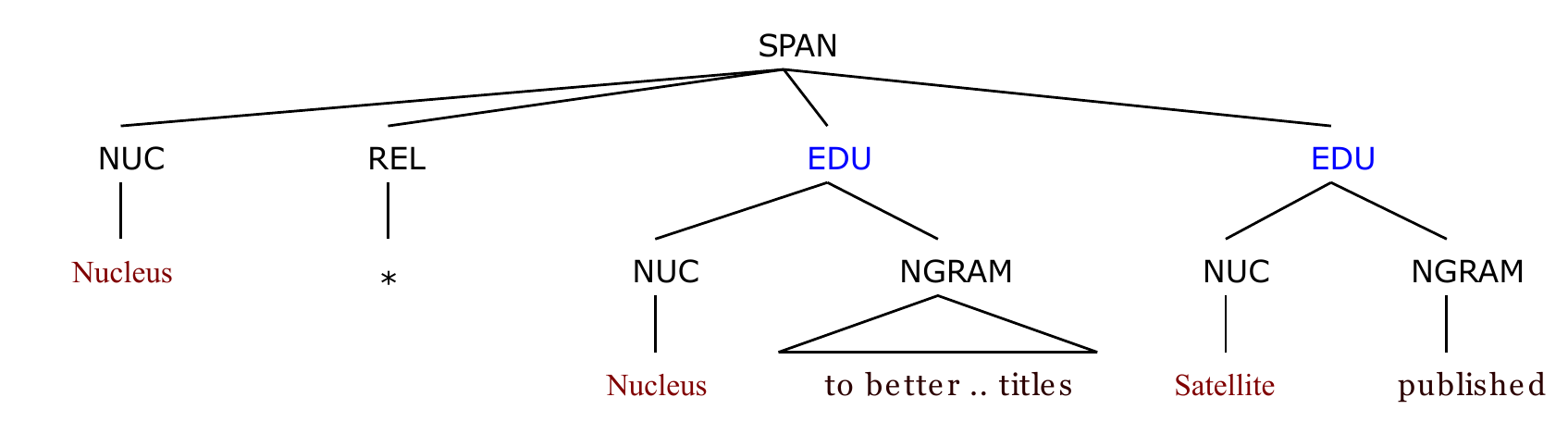} & {\small (2) \nor}\\
\includegraphics[width=.7\textwidth]{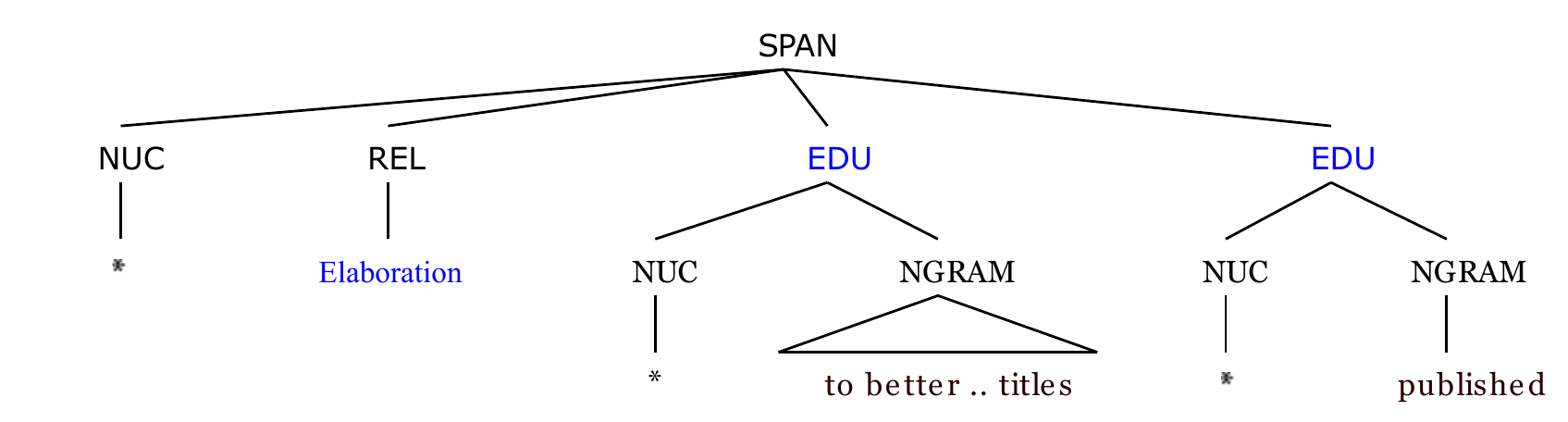} & {\small (3) \non}\\
\includegraphics[width=.7\textwidth]{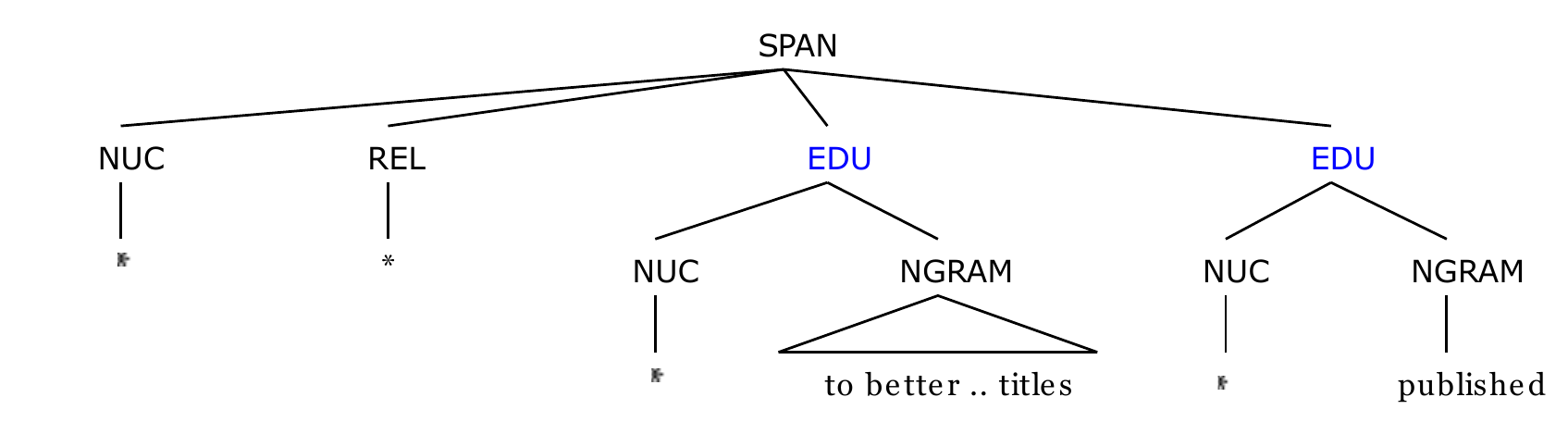}& {\small (4) \nonr}\\
\includegraphics[width=.22\textwidth]{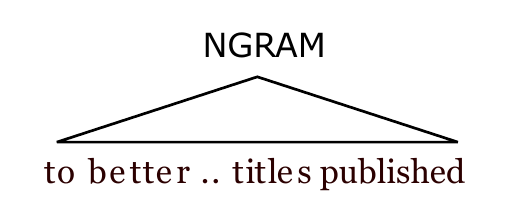}& {\small (5) \nostr}
\end{tabular}
\caption{\label{fig:ablation} Different discourse trees for the same example translation: "to better link the creation of content for all the titles published", with decreasing amount of discourse information. The five representations correspond to the ones used in the ablation study.}
\end{figure}

In our ablation experiments we contrast the original \qcril\ metric, computed over full RST trees, to variations of the same, where the discourse trees have less information. When removing a particular element, we replace the corresponding labels by a \dummy tag (`$*$'). We have the following ablation conditions, which are illustrated in Figure~\ref{fig:ablation}:

\begin{enumerate}
\item \emph{Full}: Original \qcril\ metric with the full (labeled) RST tree structure.
\item \emph{No discourse relations}: We replace all relation labels (e.g., Attribution, Elaboration, etc.) in the tree by a \dummy tag.
\item \emph{No nuclearity}: We replace all the nuclearity statuses (i.e., Nucleus, Satellite), by a \dummy tag.
\item \emph{No relation and no nuclearity tags}: We replace both the relation and the nuclearity labels by dummy tags. This leaves the discourse structure (i.e., the skeleton of the tree) along with the lexical items.
\item \emph{No discourse structure}: We remove all the discourse structure, and we only leave the lexical information. Under this representation, the evaluation metric corresponds to unigram lexical matching.
\end{enumerate}

We scored all modified trees using the same tree kernel that we used in \qcril, and we scored their resulting rankings accordingly. The summarized system-level results for WMT11-13 are shown in Table~\ref{tab:ablation}, where we used all into-English language pairs.


\begin{table}[t]
\begin{center}
{\small\begin{tabular}{llrrrrrrrr}
& & \multicolumn{2}{c}{\bf 2011} & \multicolumn{2}{c}{\bf 2012} & \multicolumn{2}{c}{\bf 2013} & \multicolumn{2}{c}{\bf Overall}\\
\cmidrule{3-10}
&\bf RST variant & $\rho$ & $r$ & $\rho$ & $r$ & $\rho$ & $r$ & $\rho$ & $r$\\
\toprule
%
\multirow{ 5}{*}{\bf DR-LEX}
 & \full  & \bf{.848} & \bf{.860} & \bf{.876} & \bf{.912} & \bf{.920} & \bf{.919} & \bf{.881} & \bf{.897} \\
 & \nor   & .843      & .856      & .876      & .909      & .919      & .919 & .879 & .895 \\
 & \non   & .822      & .828      & .867      & .896      & .910      & .914 & .866 & .879 \\
 & \nonr  & .815      & .826      & .847      & .891      & .915      & .913 & .859 & .877 \\
 & \nostr & .794      & .798      & .865      & .863      & .887      & .903 & .849 & .855 \\
\bottomrule
\end{tabular}}
\end{center}
\caption{\label{tab:ablation} System-level Spearman ($\rho$) and Pearson ($r$) correlation results for the ablation study over the \qcril\ metric across the WMT\{11,12,13\} datasets and overall.}
\end{table}

We can see a clear pattern in Table~\ref{tab:ablation}.
Starting from the lexical matching (\nostr), each layer of discourse information helps to improve performance, even if just a little bit. Overall, we observe a cumulative gain from 0.849 to 0.881 in terms of Spearman's $\rho$. Having only the discourse structure (\nonr) improves the performance over using lexical items only. This means that identifying the boundaries of the discourse units in the translations (i.e., which lexical items corresponds to which EDU), and how those units should be linked, already can tell us something about the quality of the translation. Next, by adding nuclearity information (\nor), we observe further improvement. This means that knowing which discourse unit is the main one and which one is subordinate is helpful for assessing the quality of the translation. Finally, using the discourse structure, the nuclearity, and the relations together yields the best overall performance. The differences are not very large, but the tendency is consistent across datasets.
 
Interestingly, the nuclearity status (\nor) is more important than the type of relation (\non). Eliminating the latter yields a tiny decrease in performance, while ignoring the former causes a much larger drop.
While this might seem counter-intuitive at first (because we think that knowing the type of discourse relation is important), this can be attributed to the difficulty of discourse parsing machine translated text. As we will observe in the next section, assigning the correct relation can be a much harder problem than predicting the nuclearity statuses. Thus, parsing errors might be undermining the effectiveness of the discourse relation information. 

Table~\ref{tab:ablationlp} presents the results of the same ablation study but this time broken down per language pair. For each language pair, all years are considered (2011-2013). Overall, we observe the same pattern as in Table~\ref{tab:ablation}, namely that all layers of discourse information are helpful to improve the results, and that the nuclearity information is more important than the discourse relation types.\footnote{Note that the results of Spearman's $\rho$ for \csen\ do not follow exactly the same pattern. This instability might be due to the small number of systems for this language pair (see Table~\ref{tab:wmt-stats}).} 

However, some differences are observed depending on the language pair. For example, Spanish-English exhibits larger improvements ($\rho$ goes from $0.942$ to $.970$) than French-English ($\rho$ goes from $0.936$ to $0.943$). This is despite both language-pairs being mature in terms of the expected quality for these systems. On another axis, the German-English language pair shows much lower overall correlation compared to Spanish-English ($0.782$ vs $0.970$). This can be the effect of the inherent difficulty of this language pair due to long-distance reordering, etc. However, note that adding all the discourse layers increases $\rho$ from $.738$ to $.782$. These observations are consistent with our findings in the next subsection, where we explore the different parts of the discourse trees at a more fine-grained level.

\begin{table}[t]
\begin{center}
{\small\begin{tabular}{llrrrrrrrr}
& & \multicolumn{2}{c}{\bf \csen} & \multicolumn{2}{c}{\bf \deen} & \multicolumn{2}{c}{\bf \esen} & \multicolumn{2}{c}{\bf \fren}\\
\cmidrule{3-10}
&\bf RST variant & $\rho^*$ & $r$ & $\rho$ & $r$ & $\rho$ & $r$ & $\rho$ & $r$\\
\toprule
%
\multirow{ 5}{*}{\bf DR-LEX}
& \full & .890        & {\bf .893}  & {\bf .782}  & { \bf.840}  & {\bf .970}  & {\bf .952}   & { \bf .943}   & { \bf .940 }\\
& \nor  & .894        & .892        & .775        & .833  & .968  & .952  & .942  & .939\\
&\non  & {\bf .899}  & .885        & .739        & .802  & .958  & .949  & .935  & .925\\
&\nonr & .895        & .884        & .720        & .798  & .935  & .950  & .940  & .917\\
&\nostr  & .833      & .861        & .738        & .743  & .942  & .930  & .936  & .919\\
\bottomrule
\end{tabular}}
\end{center}
\caption{\label{tab:ablationlp} System-level Spearman ($\rho$) and Pearson ($r$) correlation results for the ablation study over the \qcril\ metric across language pairs for the WMT\{11,12,13\} datasets.}
\end{table}

\subsection{Discriminating between Good and Bad Translations}
\label{subsec:relations}
In the previous subsection, we analyzed how different parts of the discourse tree contribute to the performance of the \qcril\ metric. In this section, we take a different approach: we investigate whether the information contained in the discourse trees helps to differentiate good from bad translations. 

In order to do so, we analyze the discourse trees generated for three groups of translations: \Ni \emph{gold}, the reference translations; \Nii \emph{good}, the translations of the top-2 best (per language pair) systems; and \Niii \emph{bad}, the translations of the worst-2 (per language pair) systems. Our hypothesis is that there are characteristics in the \emph{good}-translation discourse trees that make them more similar to the \emph{gold}-translation trees than the \emph{bad}-translation trees. The characteristics we analyze here are the following: relation labels, nuclearity labels, tree depth, and number of words. We perform the  analysis at the sentence level, by comparing the trees of the gold, good, and bad translations. 


\subsubsection{Discourse Relations}

\begin{figure}
\centering
\includegraphics[width=0.95\textwidth]{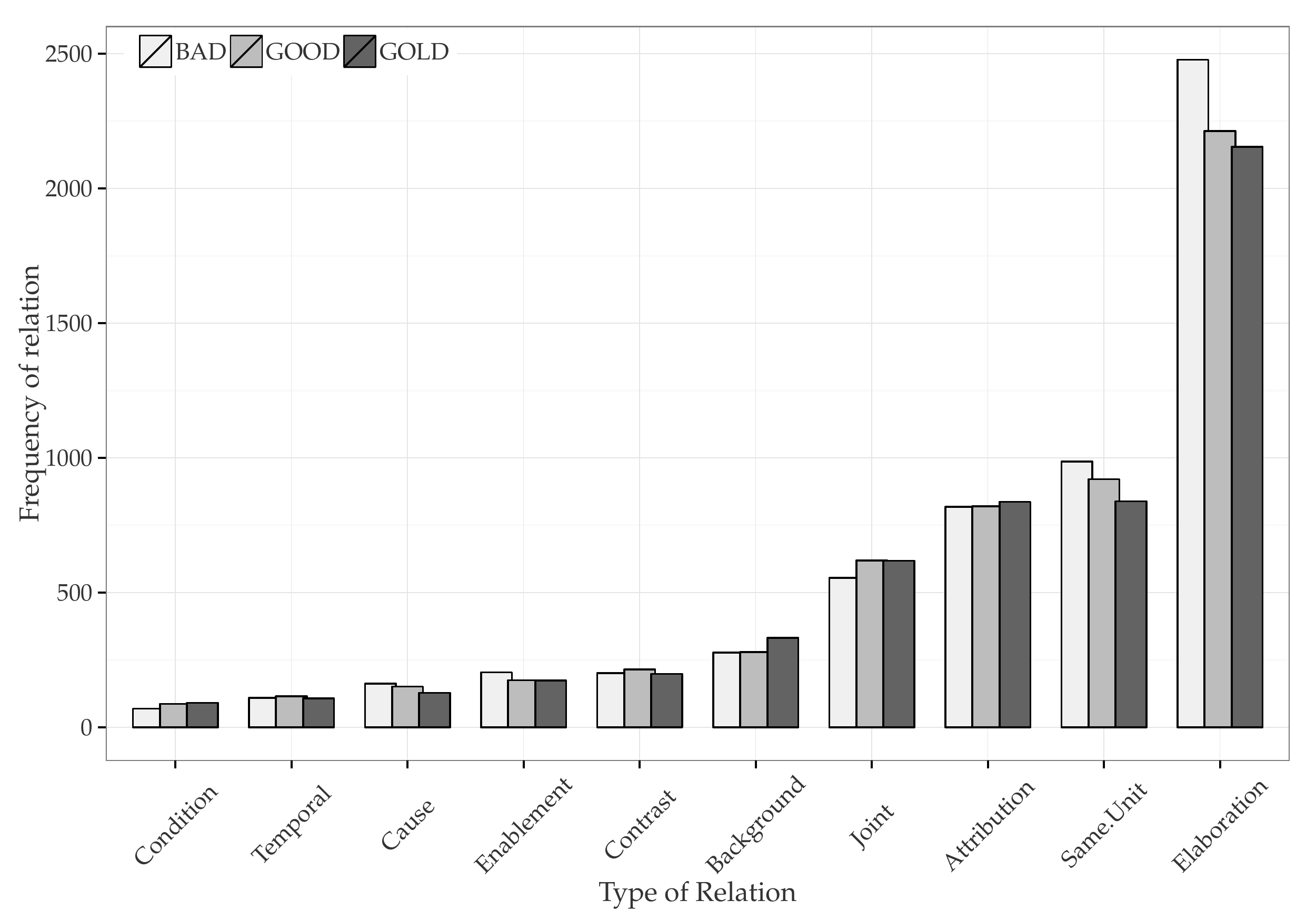}
\caption{\label{fig:rels2012}Distribution of discourse relations for \emph{gold}, \emph{good} and \emph{bad} automatic translations on WMT12 Spanish-English. We show the ten most frequent discourse relations only.}
\end{figure}

There are eighteen discourse relation labels in our RST parser. We separately computed the label frequency distributions from the RST trees of all \emph{gold}, \emph{good}, and \emph{bad} translation hypotheses. Figure~\ref{fig:rels2012} shows the histogram for the ten most frequent classes on the Spanish-English portion of the WMT12 dataset. We can see that there are clear differences between the \emph{good} and the \emph{bad} distributions, especially in the frequencies of the most common tags (\emph{Elaboration} and \emph{Same-Unit}). The \emph{good} hypotheses have a distribution that is much closer to the human references (\emph{gold}). For example, the frequency difference for the \emph{Elaboration} tag between \emph{good} and \emph{gold} translation trees is $58$, which is smaller than the difference between \emph{bad} and \emph{gold}, $323$. In other words, the trees for \emph{bad} translations exhibit a surplus of \emph{Elaboration} tags. 

If we compare the entire frequency distribution across different relations for the whole WMT12, we observe that the Kullback--Leibler (KL) divergence \cite{kullback1951information} between the \emph{good} and the \emph{gold} distributions is also smaller than the KL divergence between the \emph{bad} and \emph{gold}: $0.0021$ vs $0.0039$, and a similar tendency holds for WMT13. 
This means that \emph{good} translations have discourse trees that encode relation tags that match the \emph{gold} translation trees. This suggests that the relation tags should be an important part of the discourse metric.

In a second step, we computed the micro-averaged \fo\ score for each relation label, taking the \emph{gold} translation discourse trees as a reference. Note that computing standard parsing quality metrics that span over constituents (e.g., \fo\ score over the constituents), would require the leaves of the two trees to be the same. In our case, we work with two different translations (one \emph{gold} and one MT-generated), which makes their RST trees not directly comparable. Therefore, we apply an approximation, and we measure \fo\ score over the total number of instances of a specific tag, 
regardless of their position in the tree. Furthermore, we also consider that every instance of a predicted tag is correct if there is a corresponding tag of the same type in the \emph{gold} tree. Effectively, this makes the number of \emph{true positives} for a specific tag equal to the minimum number of instances for that tag in either the hypothesis or the \emph{gold} trees.
%
Although this is a simplification, this gives us an idea of how closely the RST trees for \emph{good}/\emph{bad} translation approximate the trees from the references.

\begin{figure}[t]
\centering
\includegraphics[width=0.95\textwidth]{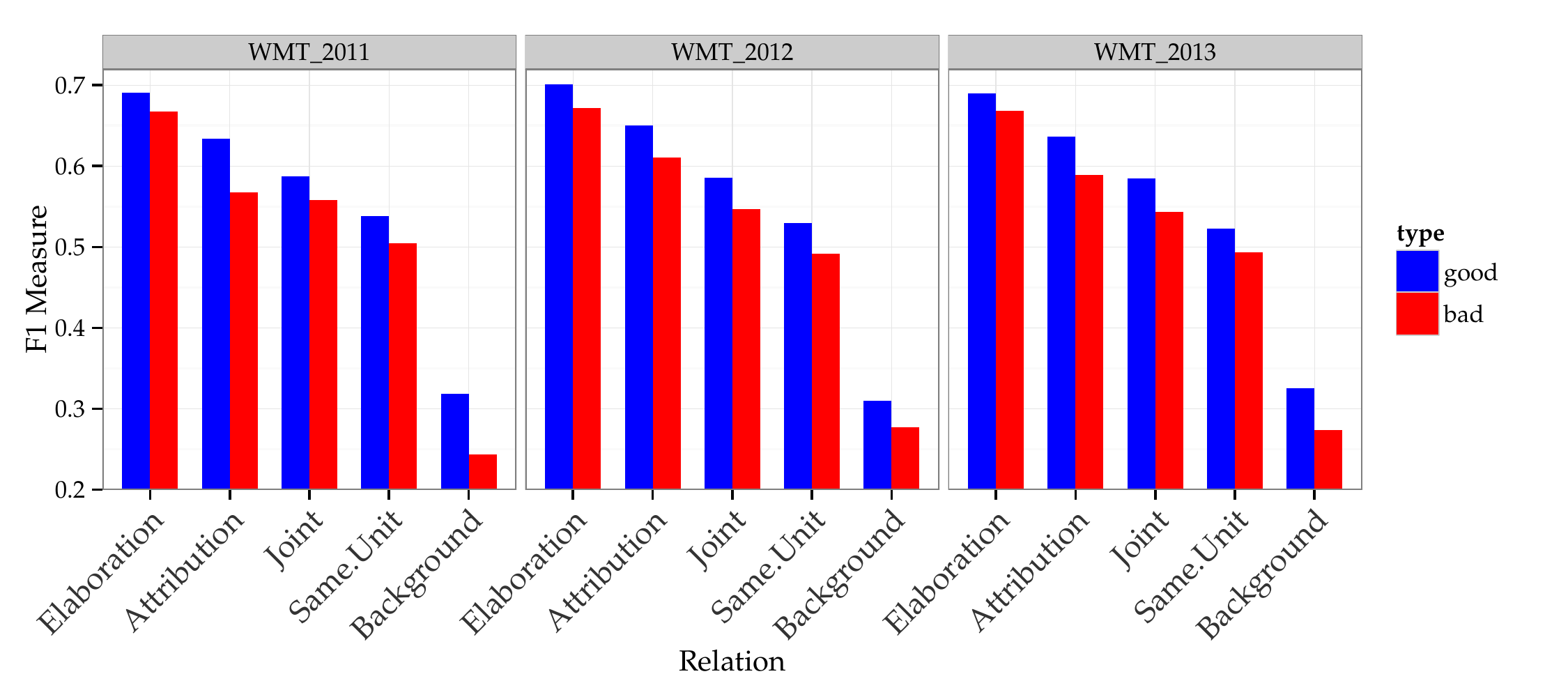}
\caption{\label{fig:relsF1} \fo\ score for each of the top-five relations in \emph{good}- vs. \emph{bad}-translation trees across the WMT\{11,12,13\} datasets.}
\end{figure}

\begin{figure}[t]
\centering
\includegraphics[width=0.95\textwidth]{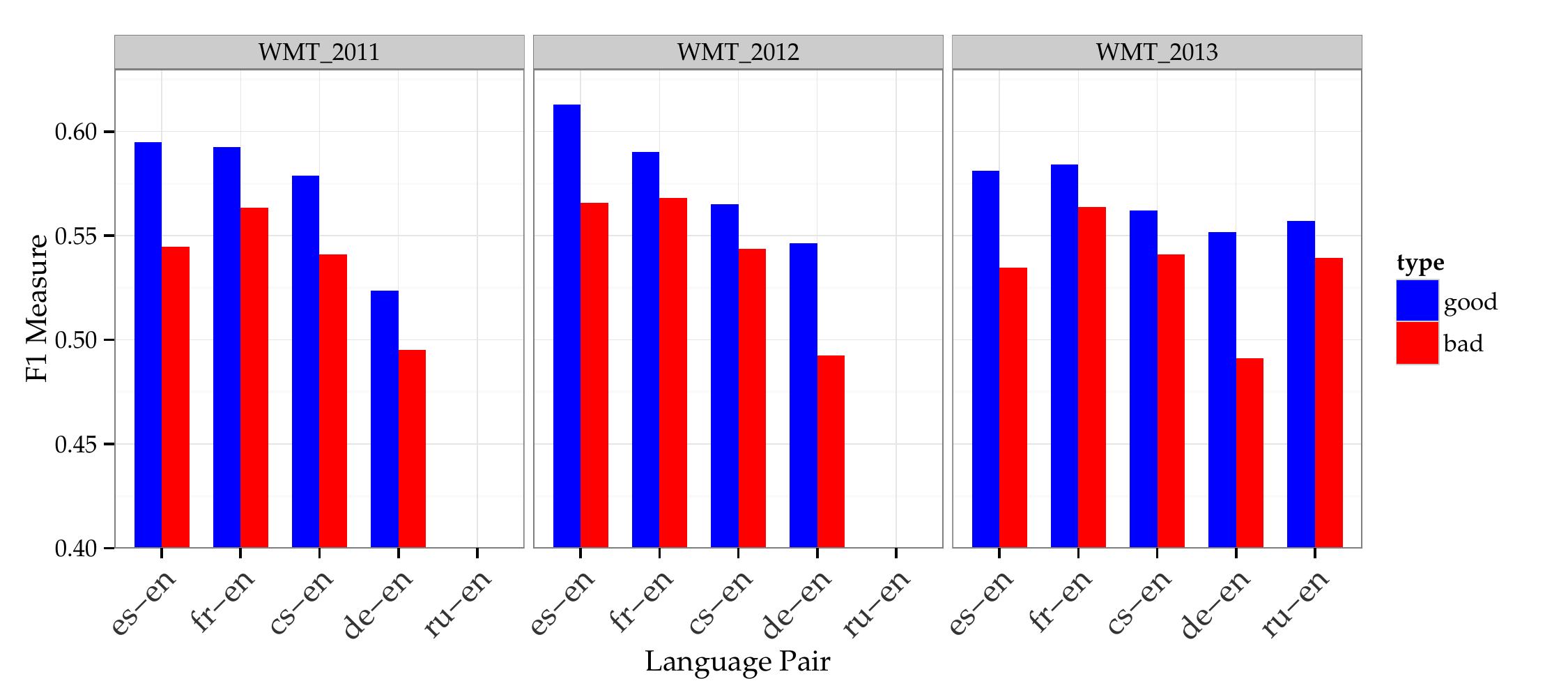}
\caption{\label{fig:relsF1lang}\fo\ score for each language pair in \emph{good}- vs. \emph{bad}-translation trees across the WMT\{11,12,13\} datasets: micro-averaging the scores of the top-five relations.}
\end{figure}

The results for the five most prevalent relations are shown in Figure~\ref{fig:relsF1}.
We can see systematically higher \fo\ scores for \emph{good} translation trees compared to \emph{bad} ones across all relations and all corpora. This supports our hypothesis, at the discourse relation level, that is, discourse trees for \emph{good} translations contain more similar discourse labels to the reference translation trees. Note, however, that \fo\ scores vary across relations and they are not very high. The highest \fo\ is around 70\%, indicating that they are hard to predict.  

Figure~\ref{fig:relsF1lang} contains the same information, but this time broken down by language pair. For each language pair and corpus year, we micro-average the results for the five most frequent discourse relations.
Again, we observe a clear advantage for the good-translation trees over the bad ones for all language pairs and for all years. Some differences are observed across language pairs, which do not always have an intuitive explanation in terms of the difficulty of the language pair.\footnote{Differences between language pairs can be attributable to the particular MT systems that participated in the competition, which is a variable that we cannot control for in these experiments.} For instance, larger gaps are observed for \esen\ and \deen, compared to the rest.
This correlates very well with the results in Table~\ref{tab:ablationlp}, clearly connecting the discourse similarity and the quality of the evaluation metrics.

\begin{figure}[t]
\centering
\includegraphics[width=\textwidth]{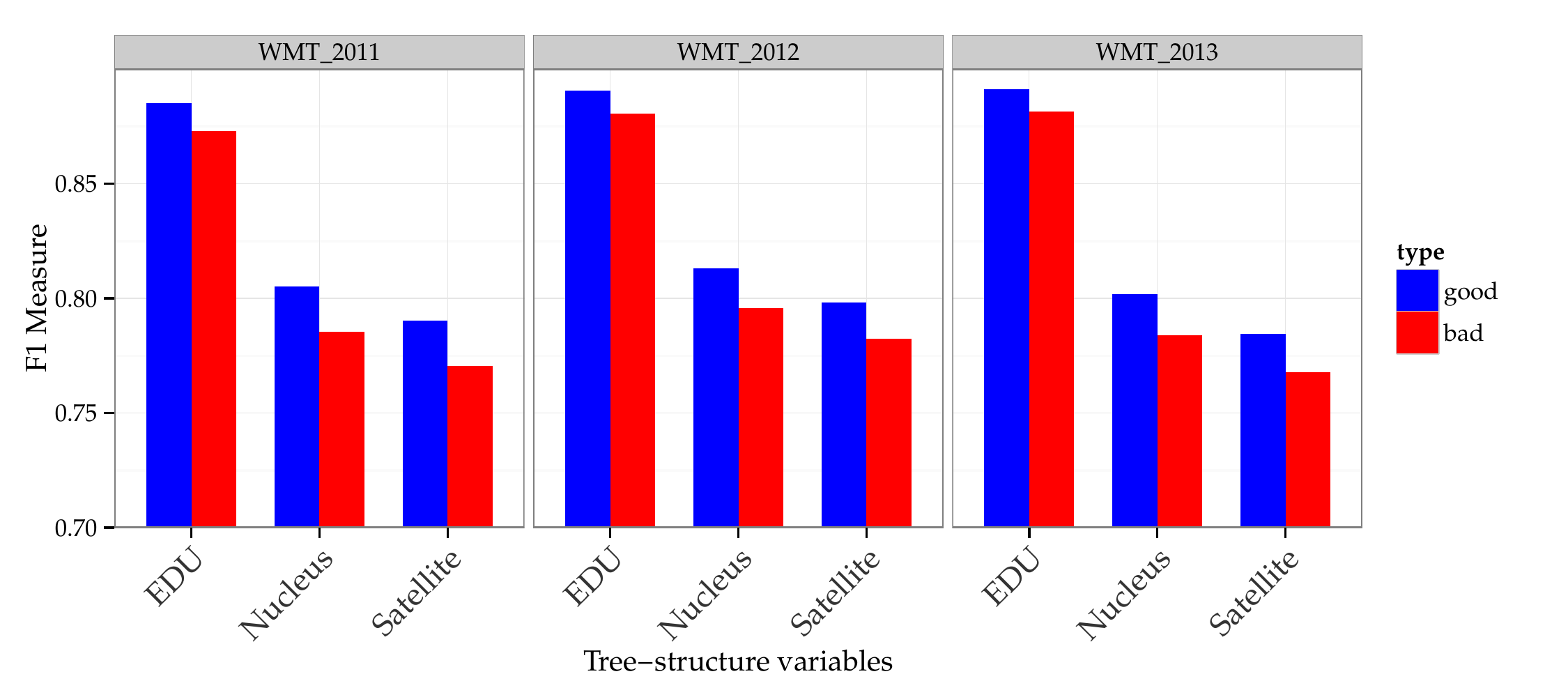}
\includegraphics[width=\textwidth]{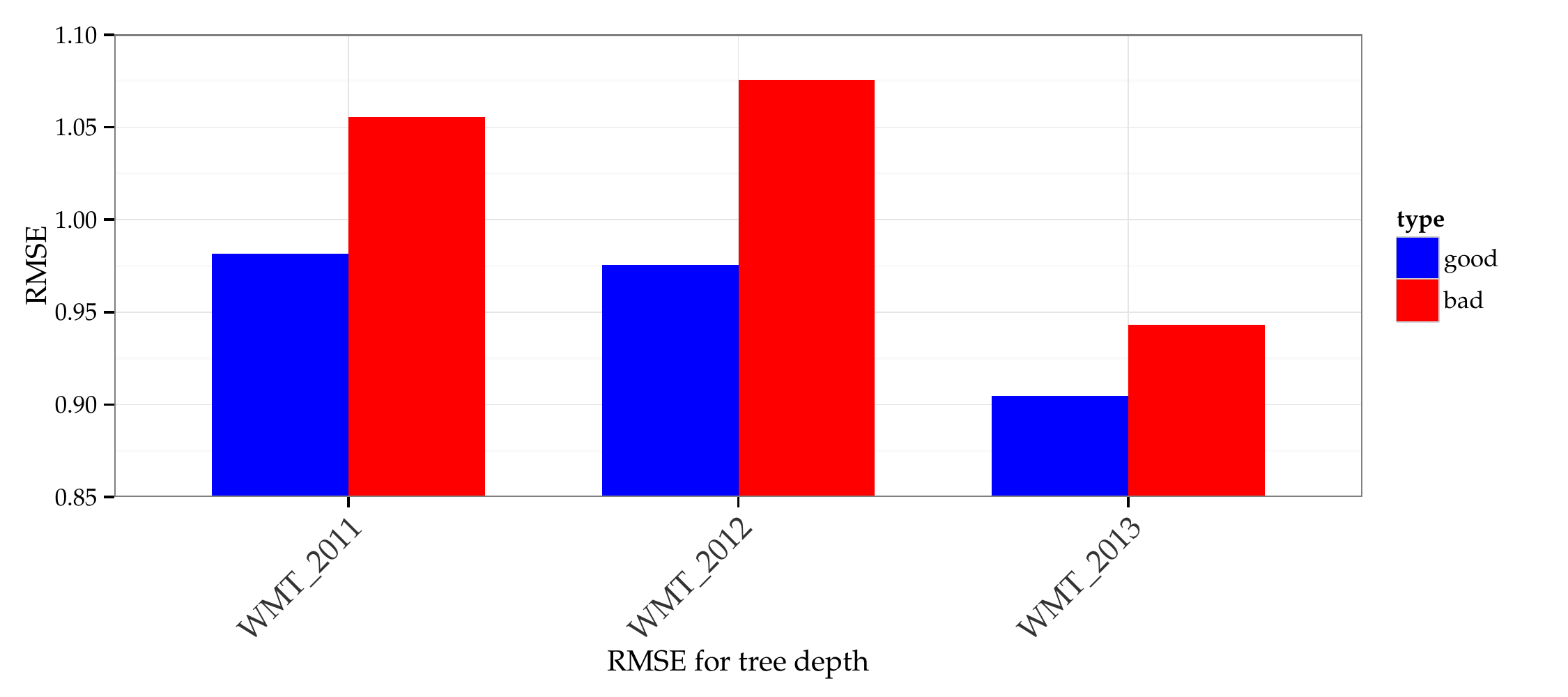}
\caption{\label{fig:nucF1}F$_1$ scores for the nuclearity relations and EDUs (upper chart), and \rmse\ for Depth (lower chart) in \emph{good}- vs. \emph{bad}-translation trees, across the WMT\{11,12,13\} datasets.}
\end{figure}

\subsubsection{Nuclearity and Other Tree Information}

\emph{Nuclearity} describes the role of the discourse unit within the relation, which can be central (\emph{Nucleus}) or supportive (\emph{Satellite}). 
Here, we study the distribution of these labels together with two extra elements from the trees: the elementary discourse units (EDUs) and the depth of the discourse tree (Depth). 
The results are shown in Figure~\ref{fig:nucF1}. For the number of \emph{Nucleus}, \emph{Satellite} and EDU labels, we  compute the simplified F$_1$ scores in the same way that we did for relation labels, focusing on the number of instances. For the tree Depth, we compute the micro-averaged root-mean-squared-error, or \rmse. 

As with the discourse relations, we observe better results for the nuclearity labels and the other tree elements from the good-translation trees, compared to the bad ones. This is consistent across all years (higher F$_1$ or lower \rmse).
Note that the \fo\ values for nuclearity labels are significantly higher than the \fo\ scores for discourse relations (now moving in the 0.78--0.82 interval, compared to \fo\ average scores below 0.60 in the case of discourse relations). This helps to explain the larger impact of the nuclearity elements in the evaluation measure (see Tables~\ref{tab:ablation} and \ref{tab:ablationlp}). Finally, predicting EDUs is even easier than predicting nuclearity labels (\fo\ values close to 0.89). Thus, EDU structure certainly contributes to improving the evaluation measure; this corresponds mainly to the \nonr\ case in the ablation study (again, see Tables~\ref{tab:ablation} and \ref{tab:ablationlp}). 




\begin{figure}[t]
\centering
\includegraphics[width=0.90\textwidth]{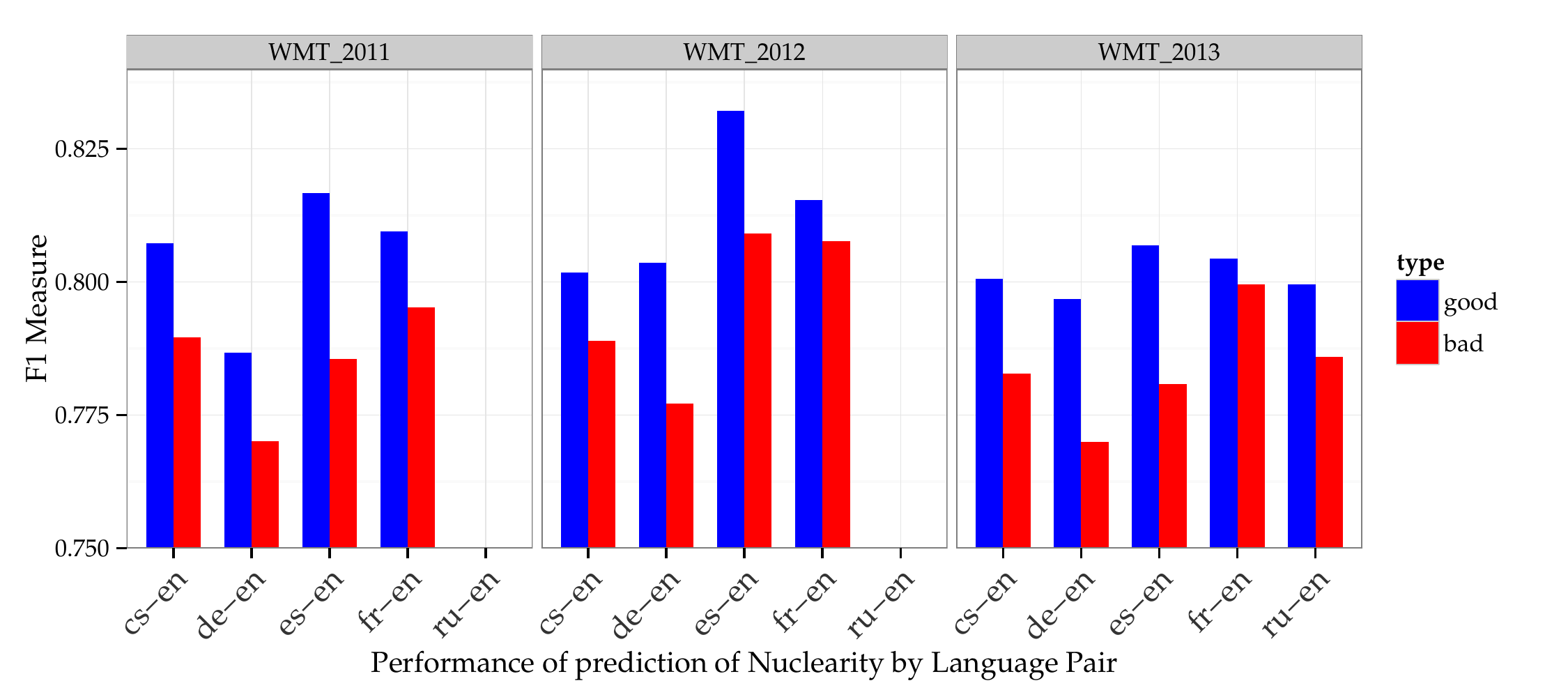}
\includegraphics[width=0.90\textwidth]{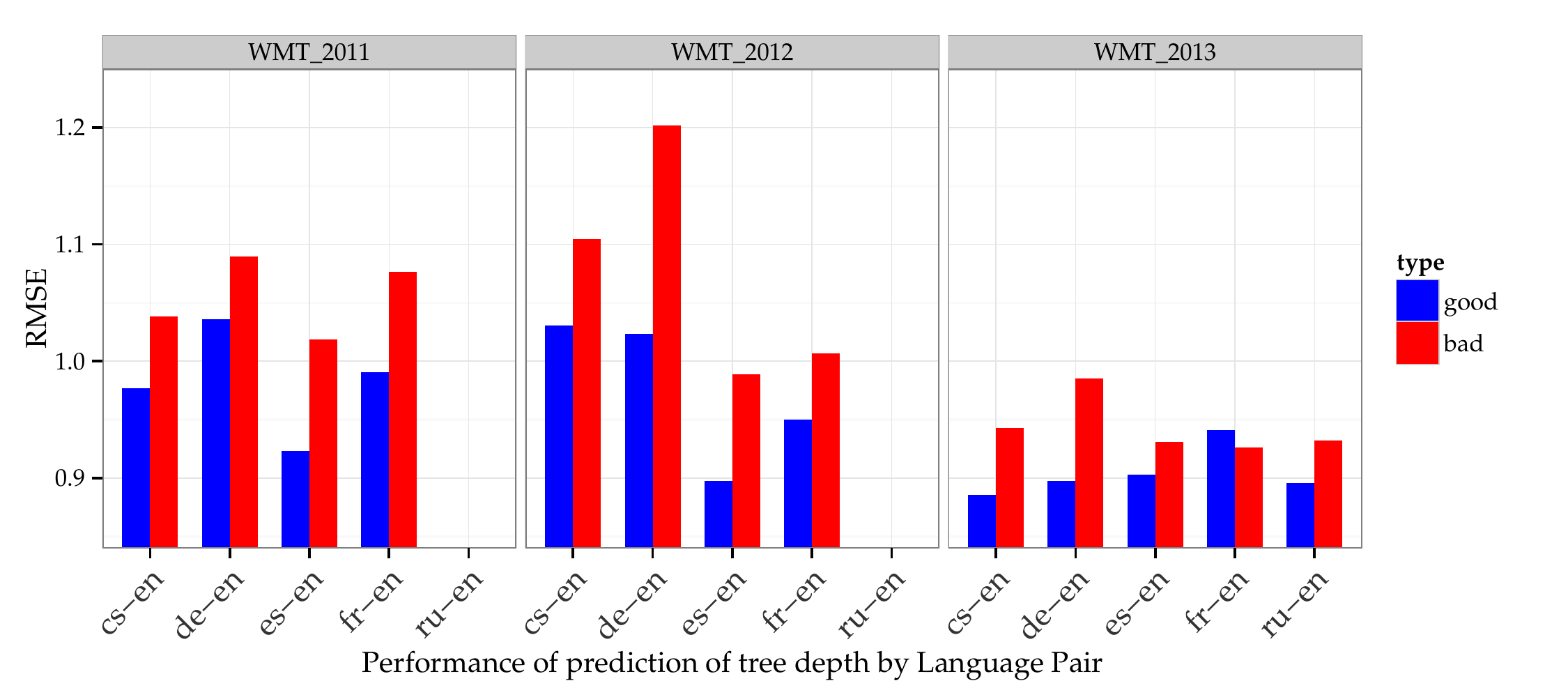}
\caption{\label{fig:nucF1lang}
Micro-averaged F$_1$ scores for each language pair for nuclearity and EDU, and \rmse\ for Depth in \emph{good}- vs. \emph{bad}-translation trees across the WMT\{11,12,13\} datasets.
}
\end{figure}

Finally, Figure~\ref{fig:nucF1lang} shows the results by language pair. 
We show the micro-averaged \fo\ scores of the nuclearity labels and EDUs (upper charts), and the \rmse\ for Depth (lower charts). Once again, the \fo\ and \rmse\ results for \emph{good} translations are better than those for \emph{bad} ones, sometimes by large margins. The only exception is for Depth in \fren\ (WMT13). 
%
Looking at the overall scores and at the size of the gaps between the scores for \emph{good} and \emph{bad}, we can see that they are consistent with the per-language results of Table~\ref{tab:ablationlp}, showing once again the direct relation between matching discourse elements and the correlation with the human assessments of the discourse-based \qcril\ metric.

The main conclusions that we can draw from this analysis can be summarized as follows: \Ni The similarity between discourse trees is a good predictor of the quality of the translation, according to the human assessments; \Nii Different levels of discourse structure and relations provide different information, which shows smooth accumulative contribution to the final correlation score; \Niii Both discourse relations and nuclearity labels have sizeable impact on the evaluation metric, the latter being more important than the former. 
The last point emphasizes the appropriateness of the RST theory as a formalism for the discourse structure of texts. Contrary to other discourse theories, e.g., the Discourse Lexicalized Tree Adjoining Grammar (D-LTAG) \cite{Webber04} used to build the Penn Discourse Treebank \cite{Prasad_pdtb_08}, RST accounts for the nuclearity as an important element of the discourse structure.


\subsection{Qualitative Analysis of Good and Bad Translations}
\label{subsec:qualitative}

In the previous two subsections we provided a quantitative analysis of which discourse information has the biggest impact on the performance of our discourse-based measure (\qcril) and also which parts of the discourse trees help in distinguishing good from bad translations. In this subsection, we present some qualitative analysis by inspecting a real example of good vs. bad translations, and showing how the discourse trees help in assigning similarity scores to distinguish them.

\begin{figure}[t]
\begin{subfigure}{\textwidth}
  \centering
  \hspace*{-4mm}\includegraphics[width=1.05\textwidth]{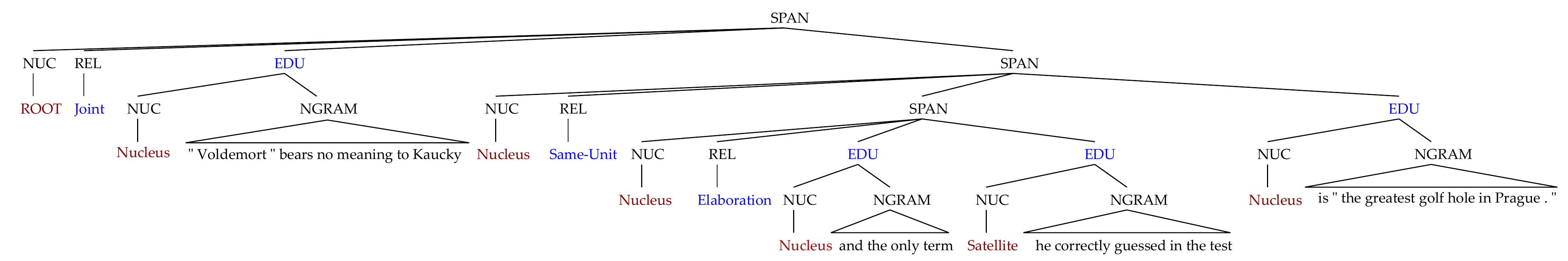}
  \caption{Reference translation\smallskip}
  \label{fig:ref_analysis}   
\end{subfigure}

\begin{subfigure}{\textwidth}
  \centering
  \hspace*{-4mm}\includegraphics[width=1.05\textwidth]{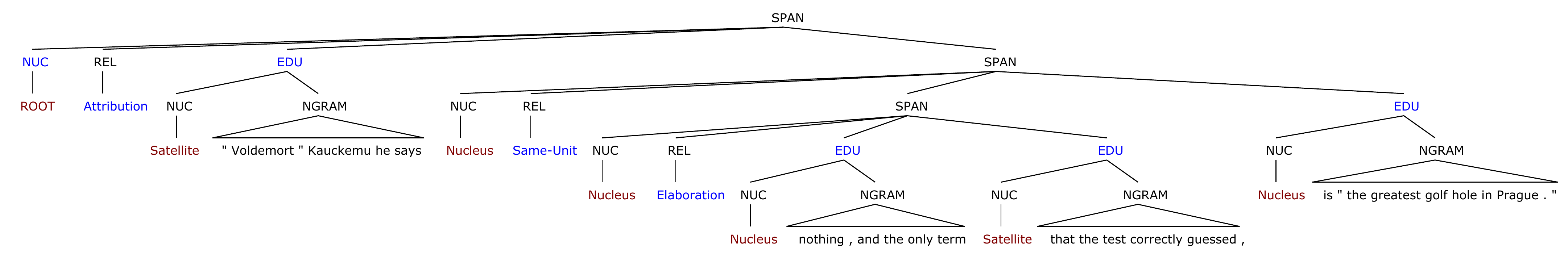}
  \caption{A \emph{better} translation with \qcril\ score $0.88$\smallskip}
  \label{fig:better}
\end{subfigure}

\begin{subfigure}{\textwidth}
  \centering
  \hspace*{-4mm}\includegraphics[width=1.05\textwidth]{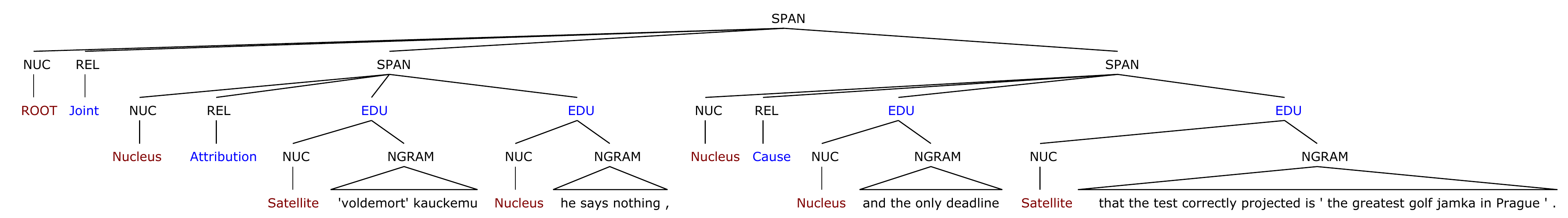}
  \caption{A \emph{worse} translation with \qcril\ score $0.75$}
  \label{fig:worse}
\end{subfigure}
\caption{Example of discourse trees for good and bad translations in comparison with the reference translation. Example extracted from WMT-2011 (\csen).}
\label{fig:good_bad_translations}
\end{figure}

Figure \ref{fig:good_bad_translations} shows a real example with discourse trees for a reference (\ref{fig:ref_analysis}) and two alternative translations, one (\ref{fig:better}) being better than the second (\ref{fig:worse}). The examples are extracted from the WMT11 dataset (\csen), and the discourse trees are obtained with our automatic discourse parser. Discourse trees are presented with the unfolded format introduced in Figure~\ref{fig:drl}.

Translation \ref{fig:better} gets a \qcril\ score of 0.88, which is higher than the score for translation \ref{fig:worse}, 0.75. Part of the difference is explained by the fact that translation \ref{fig:better} provides better word-based translation, including complete EDU constituents (e.g., \emph{is ``the greatest golf hole in Prague''}). But also, \ref{fig:better} obtains many more subtree matches with the reference at the level of the discourse structure. This translation has the same discourse structure and labels as the reference, with the only exception of the top most  discourse relation (\emph{Joint} vs. \emph{Attribution}). This tendency is observed across the datasets, and it is quantitatively verified in previous Subsection~\ref{subsec:relations}, i.e., good translations tend to share the tree structure and labels with the reference translations.

On the other hand, translation \ref{fig:worse} is much more ungrammatical. This leads to inaccurate parsing, producing a discourse tree that is flatter than the reference discourse tree and which has many more inaccuracies at the discourse relation level. 
Consequently, the tree kernel finds fewer subtree matches, and the similarity score becomes lower.

Note that the proposed kernel-based similarity assigns the same weight to all subtree matches encountered, so it is not possible for the metric to modulate which are the most important features to distinguish better from worse translations.\footnote{This is also explaining why translation \ref{fig:worse} obtains a relatively high score of 0.75. There are many subtree matches coming from the words, and from some very simple and meaningless fragments in the discourse tree (e.g., NUC--Nucleus, NUC--Satellite, etc.).} The success of the metric is based solely on the assumption (verified in~\ref{subsec:relations}) that better translations will exhibit discourse trees that are closer to the reference. A natural step to follow would be to try to learn with preference and convolutional kernels which of these subtree structures (understood as implicit features) help to discriminate better from worse translations. This is the approach followed by \namecite{guzman-EtAl:2014:EMNLP2014}, which is also mentioned in Subsection~\ref{subsec:beyondsyntax}.

\subsection{Is Discourse Providing Relevant Information beyond Syntax?}
\label{subsec:beyondsyntax}

\begin{table*}[t]
\centering
\small\begin{tabular}{lcccccc|}
  \hline
  \bf Structure & \csen & \deen & \esen & \fren & \multicolumn{1}{c}{Overall} \\
  \hline
 Syntax     & 0.190 & 0.244 & 0.198 & 0.158 & 0.198 \\
 Discourse  & 0.176 & 0.235 & 0.166 & 0.160 & 0.184 \\
 Syntax+Discourse & \bf 0.210 & \bf 0.251 & \bf 0.240 & \bf 0.223 & \bf 0.231 \\
\hline
\end{tabular}
\caption{\label{tab:syn+dis}Kendall's ($\tau$) segment level correlation with human judgements on WMT12 obtained by the pairwise preference kernel learning. 
 Results are presented for each language pair and overall.}
\vspace{-1em}
\end{table*}


Discourse parsing at the sentence level relies heavily on syntactic features extracted from the syntactic parse tree. One valid question to raise is whether the sentence-level discourse structure is providing any relevant information for MT evaluation apart from the syntactic relations. Note that in Subsection~\ref{subsec:discotk} we combined up to 18 metrics with our discourse-based evaluation metrics. Three of them use dependency parsing features (DP-HWCM-c-4, DP-HWCM-r-4, and DP-Or*; cf. Figure~\ref{fig:freq}) and a fourth one uses constituency parse trees (CP-STM-4; same figure).\footnote{\final{\asiya\ syntactic metrics are described in pages 21-24 of the 
manual (\url{http://asiya.lsi.upc.edu/}).}} 
According to the interpolation weights, the contribution of these metrics is not negligible, but it seems to be lower than that of the DR metrics. Still, this is a too indirect way of approaching the comparison.



Our previous work~\cite{guzman-EtAl:2014:EMNLP2014} helps answer the question about the complementarity of the two sources of information in a more direct way. In that paper, we proposed a pairwise setting for learning MT evaluation metrics with preference tree kernels. The setting can incorporate syntactic and discourse information encapsulated in tree-based structures and the objective is to learn to differentiate better from worse translations by using all subtree structures as implicit features.
The discourse parser we used is the same used in this paper. The syntactic tree is mainly constructed using the Illinois chunker~\cite{PunyakanokRo01}.
The kernel used for learning is a preference kernel~\cite{Shen-Joshi:2003:CONLL,DBLP:conf/cikm/Moschitti08}, which decomposes into Partial Tree Kernel~\cite{Moschitti-ECML2006} applications between pairs of enriched tree structures. Word unigram matching is also included in the kernel computation, thus being quite similar to \qcril.

Table~\ref{tab:syn+dis} shows the results obtained on the same WMT12 dataset by using only discourse structures, only syntactic structures or both structures together.
As we can see, the $\tau$ scores of the syntactic and the discourse variants are not very different (with a general advantage for syntax), but when put together there is a sizeable improvement in correlation for all the language pairs and overall. This is clear evidence that the discourse-based features are providing additional information, which is not included in syntax.

%% file: sections/related.tex
%






\final{Below we provide a brief overview of related work on discourse in MT (Subsection \ref{subsec:discoinMT}), followed by work on MT evaluation (Subsection \ref{subsec:discoinMTEval}). In the latter, we cover MT evaluation in general, and in the context of discourse analysis. We also discuss our  previous work on using discourse for MT evaluation.}


\subsection{Discourse in Machine Translation}
\label{subsec:discoinMT}

The earliest work on using discourse in machine translation that we are aware of dates back to 2000: \namecite{Marcu00_mt} proposed rewriting discourse trees for MT.
However, this research direction was largely ignored by the research community as the idea was well ahead of its time: note that it came even before the nowadays standard phrase-based SMT model was born \cite{Koehn:2003:SPT}.

Things have changed since then, and today there is a vibrant research community interested in using discourse for MT, \final{which has started its own biannual Workshop on Discourse in Machine Translation, \emph{DiscoMT} \cite{DiscoMT2013,DiscoMT2015,DiscoMT2017}. 
The 2015 edition also started a shared task on cross-lingual pronoun translation \cite{DiscoMT2015:sharedtask}, which had a continuation at WMT 2016~\cite{guillou-EtAl:2016:WMT}, and which is now being featured also at DiscoMT 2017. These shared tasks have the goals of establishing the state of the art, and creating common datasets that would help future research in this area.}

By now, several discourse-related research problems have been explored in MT: 
\begin{itemize}
\item \textbf{consistency in translation}~\cite{Carpuat2009,Carpuat:Simard:2012,Ture:2012,guillou:2013:DiscoMT};
\item \textbf{lexical and grammatical cohesion and coherence}~\cite{tiedemann:2010:WMT,tiedemann:2010:DANLP,gong-zhang-zhou:2011:EMNLP,Hardmeier12,Voigt2012,Wong2012,ben-EtAl:2013:Short,xiong-EtAl:2013:EMNLP,louis-webber:2014:EACL,tu-zhou-zong:2014:P14-1,Xiong:2015};
\item \textbf{word sense disambiguation}~\cite{Vickrey:2005,Carpuat2007,chan-ng-chiang:2007:ACLMain};
\item \textbf{anaphora resolution and pronoun translation}~\cite{Hardmeier2010,LeNagard:2010,Guillou:2012,Popescu-Belis-LREC-2012}; 
\item \textbf{handling discourse connectives}~\cite{pitler-nenkova:2009:Short,Becher:2011,Cartoni2011,Meyer2011,Meyer2012,meyer-popescubelis:2012,Hajlaoui_CAASL4-AMTA2012_2012,Popescu-Belis-LREC-2012,meyer-polakova:2013:DiscoMT,meyer-webber:2013:DiscoMT,li-carpuat-nenkova:2014,steele:2015:SRW};
\item \textbf{full discourse-enabled MT} \cite{Marcu00_mt,tu-zhou-zong:2013:Short}.
\end{itemize}

\final{\noindent More details on discourse-related research for MT can be found in the survey \cite{Hardmeier2012}, as well as in the PhD thesis \emph{Discourse in Statistical Machine Translation} 
\cite{Hardmeier2014}, which received the EAMT Best Thesis Award in 2014.}


\subsection{Discourse in Machine Translation Evaluation}
\label{subsec:discoinMTEval}

Despite the research interest, 
so far most attempts to incorporate discourse-related knowledge in MT
have been only moderately successful, at best.\footnote{A notable exception is the work of \namecite{tu-zhou-zong:2013:Short}, who report up to 2.3 BLEU points of improvement for Chinese-to-English translation using an RST-based MT framework.}
A common argument is that current automatic evaluation metrics such as \bleu \
are inadequate to capture discourse-related aspects of translation quality~\cite{Hardmeier2010,Meyer2012,Meyer_thesis_2014}.
Thus, there is consensus that discourse-informed MT evaluation metrics are needed in order to advance MT  research.

The need to consider discourse phenomena in MT evaluation was also emphasized earlier by the Framework for Machine Translation Evaluation in ISLE (FEMTI) \cite{Hovy_MT:2002}, which defines quality models (i.e., desired MT system qualities and their metrics) based on the intended context of  use.\footnote{\url{http://www.isi.edu/natural-language/mteval/}} The \emph{suitability} requirement of MT system in the FEMTI comprises discourse aspects including \emph{readability, comprehensibility, coherence, and cohesion}.

In Section~\ref{sec:evaluation} above, we have suggested some simple ways to create such metrics, and we have also shown that they yield better correlation with human judgments. Indeed, we have shown that using linguistic knowledge related to discourse structures can improve existing MT evaluation metrics. Moreover, we have further proposed a state-of-the-art evaluation metric that incorporates discourse information as one of its information sources.

Research in automatic evaluation for MT is very active, and new metrics are
constantly being proposed, especially in the context of the MT metric comparisons \cite{WMT07} 
\final{and metric shared tasks that ran as part of the Workshop on Statistical Machine Translation, or WMT, 
\cite{WMT08,WMT09,WMT10,WMT11,WMT12,machavcek-bojar:2013:WMT,machacek-bojar:2014:W14-33,stanojevic-EtAl:2015:WMT,bojar-EtAl:2016:WMT2},
and the NIST Metrics for Machine Translation Challenge, or MetricsMATR.\footnote{\url{http://www.itl.nist.gov/iad/mig/tests/metricsmatr/}}
For example, at WMT15, 11 research teams submitted 46 metrics to be compared \cite{stanojevic-EtAl:2015:WMT}.}

Many metrics at these evaluation campaigns explore ways to incorporate syntactic and semantic  knowledge. This reflects the general trend in the field.
For instance, at the syntactic level, we find metrics that measure the structural similarity between shallow syntactic sequences \cite{Gimenez2007,Popovic2007} or between constituency trees \cite{Liu2005}.
In the semantic case, there are metrics that exploit the similarity over named entities,
predicate-argument structures~\cite{Gimenez2007,Lo2012}, or semantic frames~\cite{Lo2011}.
Finally, there are metrics that combine several lexico-semantic aspects \cite{AsiyaMTjournal:2010}.
%


As we mentioned above, one problem with discourse-related MT research is that it might need specialized evaluation metrics to measure progress. This is especially true for research focusing on relatively rare discourse-specific phenomena, as getting them right or wrong might be virtually ``invisible'' to standard MT evaluation measures such as \bleu, even when manual evaluation does show improvements \cite{Meyer2012,taira-sudoh-nagata:2012:SSST-6,novak-nedoluzhko-zabokrtsky:2013:DiscoMT}.

Thus, specialized evaluation measures have been proposed
e.g., for the translation of discourse connectives \cite{Meyer2012,Hajlaoui_CAASL4-AMTA2012_2012,Hajlaoui_ACL-WMT-13_2013}
and for pronominal anaphora \cite{Hardmeier2010}, among others.

In comparison to the syntactic and semantic extensions of MT metrics,
there have been very few previous attempts to incorporate discourse information. 
One example are the semantics-aware metrics of \namecite{Gimenez2009} and \namecite{Comelles2010},
which used the Discourse Representation Theory~\cite{Kamp1993}
and tree-based discourse representation structures (DRS) produced by a semantic parser.
They calculated the similarity between the MT output and the references
based on DRS subtree matching as defined in~\cite{Liu2005},
using also DRS lexical overlap, and DRS morpho-syntactic overlap.
However, they could not improve correlation with human judgments
as evaluated on the MetricsMATR dataset, which consists of 249 manually-assessed segments.
Compared to that previous work, here
\Ni \ we used a different discourse representation (RST),
\Nii \ we compared discourse parses using \emph{all-subtree} kernels \cite{Collins01},
\Niii \ we evaluated on much larger datasets, for several language pairs and for multiple metrics,
and
\Niv \ we did demonstrate better correlation with human judgments.

Recently, other discourse-related extensions of MT metrics (such as \bleu, \ter \ and \meteor) 
were proposed \cite{WongPKW11,Wong2012}, which use document-level \emph{lexical cohesion} \cite{Halliday76}.
In that work, lexical cohesion is achieved using word repetitions
and semantically similar words such as synonyms, hypernyms, and hyponyms.
For \bleu \ and \ter, they observed improved correlation with human judgments on the MTC4 dataset (900 segments)
when linearly interpolating these metrics with their \emph{lexical cohesion} score.
However, they ignored a key property of discourse, i.e., \emph{the coherence structure}, which we have effectively exploited in both tuning and no-tuning scenarios. Furthermore, we have shown that the similarity in discourse trees can yield improvements in a larger number of existing MT evaluation metrics.
Finally, unlike their work, which measured lexical cohesion at the document-level,
here we are concerned with \emph{coherence (rhetorical) structure}, primarily at the sentence-level.


\final{Finally, we should note our own previous work, on which this journal paper is based.
In \cite{guzman-EtAl:ACL2014}, we showed that using discourse can improve a number of 
pre-existing evaluation metrics, while in \cite{discoMT:WMT2014} we presented our \disco\ family of discourse-based metrics. In particular, the \discoparty\ metric (discussed in Section~\ref{subsec:discotk}) combined several variants of a discourse tree representation with other metrics from the \asiya\ MT evaluation toolkit, and yielded the best-performing metric in the WMT14 Metrics shared task.}
Compared to those previous publications of ours, here we provide additional detail and extensive analysis,
trying to explain why discourse information is helpful for MT evaluation.

\final{In another related publication \cite{guzman-EtAl:2014:EMNLP2014}, we proposed a pairwise learning-to-rank approach to MT evaluation that learns to differentiate better from worse translations comparing to a given reference. There, we integrated several layers of linguistic information, combing POS, shallow syntax and discourse parse, which we encapsulated in a common tree-based structure.}

\final{We used 
preference re-ranking kernels to learn the features automatically. The evaluation results show that learning in the proposed framework yields better correlation with human judgments than computing the direct similarity over the same type of structures. 
Also, we showed that the structural kernel learning (SKL) can be a general framework for MT evaluation, in which syntactic and semantic information can be naturally incorporated.}

\final{Unfortunately, learning features with preference kernels is computationally very expensive, both at training and at testing time. Thus, in a subsequent work \cite{guzman-EtAl:2015:ACL-IJCNLP}, we used a pairwise \emph{neural network} instead, where lexical, syntactic and semantic information from the reference and the two hypotheses is compacted into small distributed vector representations, and fed into a multi-layer neural network that models the interaction between each of the hypotheses and the reference, as well as between the two hypotheses. 
This framework yielded correlation with human judgments that rivals the state of the art.  In future work, we plan to incorporate discourse information in this neural fraemwork, which we could not do initially due to the lack of discourse embeddings. 
However, with the availability of a neural discourse parser like the one proposed by \citet{li-li-hovy:2014:EMNLP2014}, this goal is now easily achievable.}

%% file: sections/conclusions.tex
%

We addressed the research question of whether sentence-level discourse structure can help the automatic evaluation of machine translation. To do so, we  defined several variants of a simple discourse-aware similarity metric, which use the all-subtree kernel to compute similarity between RST trees. We then used this similarity metric to automatically assess MT  quality in the 
evaluation benchmarks from the WMT metrics  shared task. We proposed to take the similarity between the discourse  trees for the hypothesis and for the reference translation, as an absolute measure  of translation quality. The results presented here can be analyzed from several perspectives:

\paragraph{\final{Applicability}} The first conclusion after a series of experimental evaluations is that the sentence-level discourse structure can be successfully leveraged to evaluate translation quality. While discourse-based metrics perform reasonably well on their own, especially at the system level, one interesting fact is that discourse information is complementary to many existing metrics for MT evaluation (e.g., \bleu, \ter, \meteor) that encompass different levels of linguistic information. At a system level, this leads to systematic improvements in correlation with human judgments in the majority 
of the cases where the discourse-based metrics were mixed with other single metrics in a uniformly weighted linear combination. When we further tuned the combination weights via supervised learning from human-assessed  pairwise examples, we obtained even better results and observed average relative gains between 22-35\% in segment-level correlations.


\paragraph{Robustness} Other interesting properties we observed in our experiments have to do with the 
robustness of the supervised learning combination approach. The results were very stable 
when training and testing across several WMT datasets from different years. Additionally,
the tuned metrics were quite insensitive to the source language in the translation, to 
the point that it was preferable to train with all training examples together rather than 
training source-language specific models. 

\paragraph{External validation} Exploiting this combination approach to its best, we produced a strong combined MT evaluation metric (\discoparty) composed by twenty three individual metrics, including five variants of our discourse-based metric, which performed best at the WMT14 translation evaluation  task, both at the system and at the segment level. When building the state-of-the-art \discoparty~metric, we observed that discourse-based features are favorably weighted (e.g., \drLEX was ranked 4th out of the twenty three metrics) having coefficients that are on par with other features such as \bleu. This tells us that the contribution of discourse-based information is significant even in the presence of such a rich diversity of sources of information. In this direction, we also presented evidence showing that the contribution from the sentence-level discourse information is beyond what the syntactic information provides to the evaluation metrics; in fact, the two linguistic dimensions collaborate well producing cumulative gains in performace.

\paragraph{Understanding the Contribution of Discourse}  In this article, we also presented a more qualitative analysis in order to better understand the contribution of the discourse trees in the new proposed discourse metrics. First, we conducted an ablation study, and we confirmed that all layers of information present in the discourse trees (i.e., hierarchical structure, discourse relations,  and nuclearity labels) play a role and make a positive incremental contribution to the final performance. Interestingly, the most relevant piece of information is the nuclearity labels, instead of the relations. 

Second, we analyzed the ability of discourse trees to discriminate between good and bad translations in practice. For that, we computed the similarity between the discourse trees of the reference translations and the discourse trees of a set of \emph{good} translations, and compared it to  the similarity between the discourse trees of the reference translations and a set of \emph{bad} translations. \emph{Good} and \emph{bad} translations were selected based on existing human evaluations. This similarity was computed at different levels, including  relation labels, nuclearity labels, elementary discourse units, tree depth, etc. We observed a systematically higher similarity to the discourse trees of \emph{good} translations, in all the specific elements tested and across all language pairs. These results confirm the ability of discourse trees to characterize \emph{good} translations as the ones more similar to the reference. 

\subsubsection*{Limitations and Future Work}
An important limitation of our study is that it is restricted to sentence-level discourse parsing. Although it is true that complex sentences with non-trivial discourse structure abound in our corpora, it is reasonable to think that there is more potential in the application of discourse parsing at the paragraph or at the document level. 
The main challenge in this direction is that there are no corpora available with manual annotations of the translation quality at the document level.

Second, we have applied discourse only for MT evaluation, but we would like to follow a similar path to verify whether discourse can also help machine translation itself. 
Our first approach will be to use discourse information to re-rank a set of candidate translations. The main challenge here is that one has to establish the links between the discourse structure of the source and that of the translated sentences, trying to promote translations that preserve discourse structure.
 
Finally, at the level of learning, we are working on how to jump from tuning the overall weights of a linear combination of metrics to perform learning on fine-grained features, e.g., consisting of the substructures that the discourse parse tree, and other linguistic structures (syntax, semantics, etc.) contain. This way, we would be learning the features that help identifying better translations compared to worse translations \cite{guzman-EtAl:2014:EMNLP2014,guzman-EtAl:2015:ACL-IJCNLP}.
Our 
vision is to have 
a model that can learn combined evaluation metrics taking into account different levels of linguistic information, fine-grained features and pre-existing measures, and which could be applied, with minor variations, to the related problems of MT evaluation, quality estimation and reranking.